\documentclass[fleqn,12pt]{wlscirep}
\usepackage{setspace} 
\usepackage{listing}

\usepackage{amsfonts}
\usepackage{mathrsfs}
\usepackage{amssymb}
\usepackage{bbm}
\usepackage{algorithm}
\usepackage[noend]{algpseudocode}
\usepackage{listings}
\usepackage{longtable}
\usepackage{pdflscape}
\usepackage{supertabular}
\usepackage{setspace}
\usepackage{verbatim}
\usepackage{wrapfig}
\usepackage{array}
\usepackage{lscape}

\usepackage{url}
\usepackage[final]{pdfpages}

\title{Guided Random Forest and its application to data approximation}

\author[1]{Prashant Gupta}
\author[1]{Aashi Jindal}
\author[1,*]{Jayadeva}
\author[2,3,4,5,*]{Debarka Sengupta}
\author[6]{Suresh Chandra}
\affil[1]{Department of Electrical Engineering, Indian Institute of Technology Delhi, Hauz Khas, Delhi 110016, India}
\affil[2]{Center for Computational Biology, Indraprastha Institute of Information Technology, Delhi 110020, India}
\affil[3]{Department of Computer Science and Engineering, Indraprastha Institute of Information Technology, Delhi 110020, India}
\affil[4]{Infosys Center for Artificial Intelligence, Indraprastha Institute of Information Technology, Delhi 110020, India}
\affil[5]{Institute of Health and Biomedical Innovation, Queensland University of Technology, Brisbane, Australia.}
\affil[6]{Department of Mathematics, Indian Institute of Technology Delhi, Hauz Khas, Delhi 110016, India}
\affil[*]{To whom correspondence should be addressed. Email: debarka@iiitd.ac.in, jayadeva@ee.iitd.ac.in\\}


\newcolumntype{L}[1]{>{\raggedright\let\newline\\\arraybackslash\hspace{0pt}}m{#1}}
\newcolumntype{C}[1]{>{\centering\let\newline\\\arraybackslash\hspace{0pt}}m{#1}}
\newcolumntype{R}[1]{>{\raggedleft\let\newline\\\arraybackslash\hspace{0pt}}m{#1}}

\usepackage{times}
\usepackage{textcomp}
\usepackage{soul}
\begin{abstract}
We present the Guided Random Forest (GRAF), an ensemble classifier that extends the idea of building oblique decision trees with localized partitioning, to obtain a global partitioning. We show that global partitioning bridges the gap between decision trees and boosting algorithms, and empirically, that it reduces the generalization error bound. Results on 115 benchmark datasets show that GRAF yields comparable or better results on a majority of datasets. We also present a new way of approximating datasets in the framework of random forests.
\end{abstract}

\begin{document}

\flushbottom
\maketitle
\thispagestyle{empty}

\section{Introduction}

In supervised learning, one aims to learn a classifier that generalizes well on unknown samples ~\cite{dietterich2000ensemble}. As commonly understood, a classifier should have an error rate better than a random guess. If a classifier has a slightly better performance than a coin toss, it is termed a weak classifier. In ensemble learning, several weak classifiers are trained, and during prediction, their decisions are combined to generate a weighted or unweighted (voting) prediction for test samples. The motivation is that the classifiers' errors are uncorrelated; hence, the combined error rate is much lower than individual ones~\cite{breiman2001random}.


It has been shown that an ensemble of trees works best as a general-purpose classifier~\cite{fernandez2014we}. Amongst several known methods for constructing ensembles, \textit{Bagging} and \textit{Boosting} are widely used. For every tree, bagging generates a new subset of training examples~\cite{breiman2001random}. Boosting assigns higher weights to misclassified samples while building an instance of a tree~\cite{friedman2001greedy,chen2016xgboost}. With either strategy, a tree in an ensemble is constructed by a recursive split of the data into two parts at every node. The split can be axis-aligned, in which the split is based on a feature ~\cite{breiman2001random,geurts2006extremely}, or oblique, where a combination of features is used~\cite{murthy1993oc1,murthy1994system} for every split.


Axis-aligned trees perform well with redundant features~\cite{menze2011oblique,wickramarachchi2016hhcart}, while oblique splits yield shallower trees~\cite{zhang2017robust}. However, memory and computational requirements are higher for oblique trees. Hence, the literature focuses on finding better splits to create shallower oblique trees. Shallower trees tend to generalize better.

Even with these limitations, oblique trees have been widely used in diverse tasks across various domains. Do \textit{et al.}~\cite{do2015classifying} apply oblique trees to fingerprint dataset classification. Qiu \textit{et al.}~\cite{qiu2017oblique} used them for time-series forecasting, Zhang \textit{et al.}~\cite{zhang2017robust} for visual tracking, and Correia and Schwartz~\cite{correia2016oblique} for pedestrian detection.


In this work, we propose Guided Random Forest (GRAF), that extends the outlook of a plane generated for a certain region to other regions as well. GRAF iteratively draws random hyperplanes, and corrects each impure region, in order to increase the purity values of resultant regions. Unlike other methods, a hyperplane in GRAF is not constrained to the region it is generated for, but is shared across all possible regions. The sharing of planes across regions reduces the number of separating hyperplanes in trees, which in turn, reduces the memory requirement. 

The resultant regions (or leaf nodes) in GRAF are represented with variable length codes. This process of tree construction bridges the gap between boosting and decision trees, where every tree represents a high variance instance. We show that GRAF outperforms state-of-the-art bagging and boosting based algorithms, like Random Forest~\cite{breiman2001random} and Gradient Boosting~\cite{friedman2001greedy}, on several datasets.

\begin{figure*}[!ht]
    \centering
\includegraphics[width=1.1\linewidth,height=2.5in]{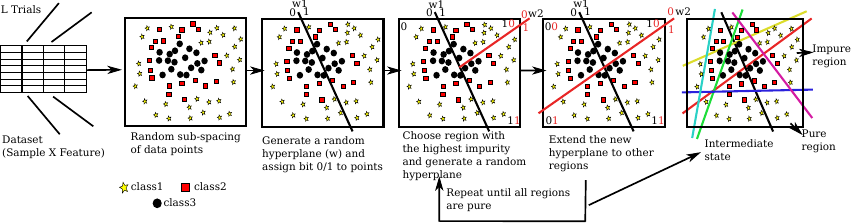}
    \caption{An overview of the creation of high variance instances in GRAF. Every instance consists of sub-spacing the dataset in a uniformly sampled feature space. A random hyperplane is generated for the sub-spaced samples. It assigns a bit 0/1 to every sample. A pure (impure) region is a region containing all (some) samples of the same class. Amongst these regions, the most impure region affects the generation of the next hyperplane. This hyperplane is extended to the other region as well, if it improves the purity of subsequent regions in that space. This generation of hyperplanes is continued until all regions are maximally purified. At an intermediate stage, regions are either pure or impure. To increase the confidence of classification, the above process is repeated to create $L$ high variance instances.}
    \label{ga}
\end{figure*}

We show that tree-based ensemble classifiers can be used for data approximation. In GRAF, the count of all random hyperplanes generated until a sample falls into its pure region, is used to assign a sensitivity value to the given sample. This assigns higher values of sensitivity to samples that lie in high confusion regions. We show that the sub-sampling of the dataset based on sensitivity scores may well approximate the entire dataset. Figure~\ref{ga} gives an overview of GRAF.

The rest of the paper is organized as follows: Section~\ref{sec::relatedWork} presents a discussion on related work; Section~\ref{section::GRAF} gives
details about GRAF; Section~\ref{section::implementation} provides the implementation details; Section~\ref{sec:graf_adaboost} explains the relationship between GRAF and boosting; Section~\ref{section::simulation} performs a simulation study to assess and compare design aspects of GRAF; Section~\ref{section::results} studies bias-variance trends and compares performances of methods on 115 UCI datasets; approximation of data using their sensitivity scores is studied in Section~\ref{section::sensitivity}; Section~\ref{section::conclusion} contains concluding remarks.

\section{Related Work}\label{sec::relatedWork}
The construction of tree-based classifiers has been an active area of research. The classifiers may differ from each other by the number of trees that are being generated, single decision tree~\cite{breiman1984classification} vs forest algorithms~\cite{breiman2001random}, or by the type of splits on nodes of the tree, axis-aligned splits~\cite{breiman2001random,geurts2006extremely} vs oblique splits~\cite{murthy1993oc1,murthy1994system}. The tree-based algorithms also differ from each other based on their size, fixed size~\cite{bennett1998support} vs top-to-down built, and the error correction methodology, misclassification~\cite{martinelli1990pyramidal,deb2002binary} vs residual error correction~\cite{friedman2001greedy,chen2016xgboost}. Amongst all these criteria, the type of split on a node has attracted a lot of attention. Two notable methods for axis-aligned splits are Random Forest (RF)~\cite{breiman2001random}, and Extremely Randomized Trees (ET)~\cite{geurts2006extremely}. RF searches for the best split using uniformly spread thresholds in the range of every feature on a node. ET increases randomness in trees and uses random thresholds for every feature. Oblique decision trees (OTs) generate splits that are not aligned with the feature axes. Since OTs consider multiple features at a time, the search space increases exponentially, doing an exhaustive search to find the best optimal oblique split impractical. Researchers have used many approximations, greedy or optimization-based, to select the best possible split. Thus, many oblique tree variants have been proposed that differ from each other in the generation of separating hyperplanes to create splits. In this section, we discuss some selected methods to generate the oblique splits.

Murthy \textit{et al.}~\cite{murthy1993oc1,murthy1994system} have proposed a decision tree, Oblique Classifier 1 (OC1), which refines the strategy of the best split selection of \textit{Classification and Regression Trees} (CART)~\cite{breiman1984classification}. OC1 employs a combination of axis-aligned and oblique splits~\cite{murthy1994system}. On a node of the decision tree, OC1 first chooses the best axis-aligned split and then looks for an oblique split. The oblique split is first generated randomly and then perturbed (one feature at a time). If perturbation shows improvement over the previously selected split, then it is kept. This process is repeated until convergence. This perturbation based optimization is prone to get stuck at a local minimum. This can be avoided by moving the converged split direction towards a random direction or restarting the perturbation step with a different initialization. Once the tree is fully grown, the OC1 tree is pruned to control over-fitting to the data. Thus, on every node, OC1 spends a significant amount of time selecting the best split. For medium-size datasets, these steps will require a significant amount of time to generate the tree. Although OC1 employs multiple heuristics to generate the split, the induced decision boundary will still be very rough, prone to poor generalization. An alternative is to construct a forest of OC1 decision trees. However, it can be argued that forest construction with simpler decision trees will be more computationally beneficial without suffering a significant loss in performance~\cite{breiman2001random}.


Tan and Dowe~\cite{tan2004mml,tan2006decision} have proposed to select an oblique split for a node based on Maximum Message Length (MML)~\cite{wallace1968information} criterion. Traditionally, MML has been used for model selection in machine learning literature~\cite{wallace1968information,wallace1999minimum,wallace2005statistical}. MML based oblique trees are generated in two steps. In the first step, the authors propose to first generate a random 2-dimensional hyperplane (oblique split) and then incrementally rotating it to generate multiple orientations. Such planes are constructed for every pair of features. For every pair, orientations with the shortest MML is selected as a possible hyperplane for the split. Amongst these hyperplanes, the ones with the smallest MML are selected as final candidate hyperplanes. The tree is grown further down, tentatively, by considering all the selected hyperplanes on the node. In the second step, a forest is created by randomly selecting one hyperplane from the final candidate hyperplanes at every node of the tree. Notice that if a dataset has $n$ features then, the total number of searches on a node is of the order $\mathcal{O}(n^2)$. Thus, for a dataset with a moderate number of features, considering all possible pairs of features may be very time-consuming. To mitigate this issue, the authors suggest limiting the maximum pairs to be searched. Although the authors experimented with only 2 or 3-dimensional split~\cite{tan2004mml}, it might be desirable to explore high dimensional splits to find further dependencies amongst features. The generalization of this method to higher dimensional splits, say $D$ ($>$2), will increase the search space of the rotated hyperplane by $(D-1)$. This makes search increasingly prohibitive with increasing $D$. Although an alternative of this is to select a few planes randomly from the set of rotated planes, a better strategy would be to search the best split among the random splits with MML criterion.




Bennet and Blue~\cite{bennett1998support} have proposed a Support Vector Machine (SVM) based formulation, called Global Tree Optimization - SVM (GTO/SVM), to induce decision trees. The proposed formulation of GTO/SVM is non-convex, and authors use \textit{hybrid extreme point tabu search} (HEPTS)~\cite{blue1998hybrid} to obtain an approximate solution. The major drawback of GTO/SVM formulation is that it requires a predefined structure of the tree. In later studies, Takahashi and Abe~\cite{takahashi2002decision} proposed a top-to-down approach to learn decision trees with SVMs. Since SVMs can handle two classes at a time, the authors presented 4 heuristics to handle multiple classes while growing trees. In the first heuristic, the authors suggested defining a binary classification problem on a node of the tree by considering one group as the class with the farthest centroid from the centroids of other classes and the remaining classes in the other group. In the second heuristic, two nearest classes are merged until only two groups are left. These steps are applied recursively to grow the decision tree. The Euclidean distance between two centroids is computed. The other two heuristics use Mahalanobis distances to create the nearest neighbor classifier and get the misclassifications of one class into another. While one heuristic tries to separate the class with the least overall misclassification error with other classes, the other heuristic merges classes with the largest misclassifications until only two groups are left. These steps are applied recursively to induce decision trees. Wang \textit{et al}~\cite{wang2006improved} proposed alternative grouping criteria based on the separability of classes. The class with the highest separability is considered one group, and other classes are grouped together to generate the split on the node. The top-to-down approach to learn decision trees with SVM has an added advantage that with a suitable kernel, non-linear decision boundaries can also be learned at every node. It, in turn, makes decision trees smaller. However, for a large dataset, the cost to store the kernels is extremely high. Also, the nodes near the leaves in the tree tend to have very few samples, making predictions unstable. Creating a forest from these trees might be one solution to mitigate this issue, but trees generated with SVM will be more or less deterministic; thus, forest generated with these decision trees will be highly correlated. Although feature sub-spacing~\cite{ho1998random} or bagging can be employed to make trees less correlated to generate trees, it will further increase the trees' storage cost. The storage issue can be mitigated by using variants of SVM with kernel approximation, such as proximal SVM. Manwani and Sastry~\cite{manwani2011geometric} suggest an alternative based on a variant of proximal SVM, Proximal SVM with Generalized Eigenvalues (GEPSVM)~\cite{mangasarian2005multisurface}. The authors argue that it is important to capture the data's geometric properties for the split criterion at each non-leaf node. To do so, the authors identify two hyperplanes, one for the majority class and the other for the remaining points. Thus, it transforms a multi-class problem also into a binary one. The remaining points are assumed to represent the other class. These hyperplanes are referred to as \textit{clustering hyperplanes}. A clustering hyperplane is closest to one class and is farthest from the patterns of the other class. Then they find two angle bisectors between the clustering hyperplanes. The angle bisector is selected based on an impurity measure, Gini impurity, as the hyperplane for that node. Zhang \textit{et al.}~\cite{zhang2014oblique,zhang2017robust} also used multi-surface proximal SVM (MPSVM) to grow decision trees. However, the authors do it differently for multi-class problems. Instead of generating only two hyperplanes for multi-class, each class is divided into two hyper-classes based on their separability.

In other studies, \textit{Rotation Forests}~\cite{rodriguez2006rotation,kuncheva2007experimental} used principal components of high variance to obtain the direction of split. Rotation Forest splits the given feature set into $k$ subsets and runs PCA separately on each subset. Thus, different splits of the feature lead to more diverse classifiers. Unlike Rotation Forests, which use unsupervised methods to obtain the split, Menze \textit{et al.}~\cite{menze2011oblique} proposed using supervised methods. They experimented with two models, one with Linear Discriminant Analysis (LDA) like projections, and another with ridge regression to obtain the split. However, with supervised methods, trees lose their inherent property of facilitating multi-class classification.


Continuously Optimized Oblique (CO2) Forest~\cite{norouzi2015co2,norouzi2015efficient} optimizes a objective function based on latent variable Support Vector Machine~\cite{yu2009learning} to select an oblique split. The objective function employed by CO2 is non-convex. To optimize the objective function, the author utilizes the convex-concave procedure~\cite{yuille2003concave}, a gradient-based optimization technique, which is solved on every node. A recently proposed Weighted Oblique Decision Trees (WO DT)~\cite{yang2019weighted} optimizes the splitting criteria on every node by considering sigmoid weights on the sample assigned to the child nodes. For optimization, L-BFGS, a gradient-based optimization technique was used. In the other study, Katuwal \textit{et al.}~\cite{katuwal2020heterogeneous} suggest selecting the splitting criteria using different kinds of linear classifiers viz. SVM, MPSVM, LDA, etc. on every node. This gives heterogeneous nature to the OTs.

In all the above-mentioned methods, for every new split, correction is limited to the region for which the split has been generated.  To the best of our knowledge, GRAF is the first attempt to explicitly extend the plane to share it with other nodes.

The tree-based algorithms have also been used in other areas such as Nearest Neighbor Search~\cite{liu2006new,dasgupta2008random}, outlier/anomaly detection~\cite{liu2008isolation}, etc. There has also been some attempt to integrate neural networks with trees. Notably, Kontschieder~\cite{kontschieder2015deep} has proposed to optimize a neural network for every node in a tree. In another effort, Katuwal \textit{et al.}~\cite{katuwal2018ensemble,katuwal2018enhancing} has proposed to combine Random Vector Functional Link Network (RVFL) with trees to create an ensemble. Neural Oblivious Decision Ensembles (NODE)~\cite{popov2019neural} is a very recently proposed deep architecture for tabular datasets.


\section{Guided Random Forest (GRAF)}\label{section::GRAF}

Let $\mathbb{R}^n$ denote the n-dimensional Euclidean space. Let $X \subseteq \mathbb{R}^n$ denote the input space, and let $Y$ denote the labels corresponding to a set of $C$ classes $\{1,..,C\}$. Let a set $S$ contain $N$ samples drawn from a population characterized by a probability distribution function $D$ over $X \times Y$. Thus the given dataset is

\begin{align}\label{eq::sampleSpace}
S = \{(x^{(i)},y_i): x^{(i)}\in X, y_i \in Y, (i=1,2,..,N)\}.
\end{align}

Let us assume that $T$ high variance classifier instances are constructed on the dataset $S$. The training of an instance involves the introduction of random hyperplanes in a forward stage-wise fashion. At a given step, a combination of these hyperplanes divides $S$ into a finite number (say $P$) of disjoint regions whose union is $S$. To be specific, a single hyperplane classifier will divide $S$ into two disjoint regions (say $\Omega_1$ and $\Omega_2$), and a combination of $d$ hyperplane classifiers will divide $S$ into at most $2^d$ regions. Let the $p$th region ($1\leq p\leq P$) be denoted by $\Omega_p$. Thus, $S=\cup_{p=1}^{P}\Omega_p$ and $\Omega_i\cap\Omega_j=\emptyset \text{ for } i\neq j$. Let $n_p$ denote the number of samples in the region $\Omega_p$. Obviously $n_p>0$, otherwise $\Omega_p$ will be an empty region and hence, have no contribution.

For each sample in the region $\Omega_p$, we generate a bit '0' or '1' such that the weights $w^{(p)} = (w_1^{(p)},...,w_n^{(p)}) \in \mathbb{R}^n$ and the bias $b^{(p)}\in \mathbb{R}$ dichotomizes the region $\Omega_p$. This is achieved by using a mapping $\lambda_p:X \to \{0, 1\}$ such that for the sample point $(x^{(i)},y_i)$ in $\Omega_p$,

\begin{align}\label{eq::bitAssignment}
\lambda_{p}(x^{(i)}) = \mathbbm{1}{\left(\sum_{j=1}^{n}(w_{j}^{(p)}x^{(i)}_{j}) + bias^{(p)} > 0\right)}.
\end{align}

Here $\mathbbm{1}(.)$ denotes the indicator function and $x^{(i)}_j$ is the $j$th component of the vector $x^{(i)}$.

We now introduce the following notations for $j=1,2,..,n$.

\begin{align}\label{eq::minEquation}
m_{j}^{(p)} = \min_{1\leq i \leq n_{p}}(x^{(i)}_j\: :\: (x^{(i)}, y_i) \in \Omega_p),
\end{align}
\begin{align}\label{eq::maxEquation}
M_{j}^{(p)} = \max_{1\leq i \leq n_{p}}(x^{(i)}_j\: :\: (x^{(i)}, y_i) \in \Omega_p),
\end{align}
\begin{align}\label{eq::meanEquation}
\mu_{j}^{(p)} = \frac{1}{n_{p}}(\sum_{i=1}^{n_{p}}x^{(i)}_j,\: (x^{(i)}, y_i) \in \Omega_p),
\end{align}
\begin{align}\label{eq::weightGeneration}
w_{j}^{(p)} \sim U(m_{j}^{(p)}+\varepsilon, M_{j}^{(p)}-\varepsilon),
\end{align}

where (\ref{eq::minEquation}), (\ref{eq::maxEquation}), and (\ref{eq::meanEquation}) represents the minimum value, maximum value, and mean value of a feature $j$ in the region $p$, respectively. Then we define bias as

\begin{align}\label{eq::biasP}
bias^{(p)} = -\sum_{j}w_{j}^{(p)}\mu_{j}^{(p)},
\end{align}
where $U(a, b)$ denotes the uniform distribution of a random variable over the interval [a,b].

The mapping $\lambda_p:X \to \{0, 1\}$ as defined at (\ref{eq::bitAssignment}) above assigns a code comprising of 0s and 1s for every sample in $\Omega_p$. A region $\Omega_p$ is said to be \textbf{pure} if it contains samples of the same class, or if samples from different classes can not be separated further. On the other hand, the region $\Omega_p$ is said to be \textbf{impure} if it contains samples of different classes, that can be further dichotomized by the addition of new hyperplanes (Figure~\ref{ga}).

Let $\mathcal{F} = \{\Omega_1,\Omega_2,..,\Omega_P\}$. We now introduce a mapping $Z:\mathcal{F} \to \mathbb{R}$ such that for $1\leq p \leq P$,

\begin{align}\label{eq::labeldef}
Z(\Omega_p) = \left(1 - \sum_{c=1}^{C}(\frac{n_{p_c}}{N_c})^2 \times (\sum_{c=1}^{C}\frac{n_{p_c}}{N_c})^{-2}\right) \times n_{p},
\end{align}

where $N_c$ denotes the total samples of class $c$, and $n_{p_c}$ denotes the samples of class $c$ in region $\Omega_p$.

The function $Z$ as defined at (\ref{eq::labeldef}) is the weighted Gini impurity function whose value $Z(\Omega_p)$ quantifies the impurity associated with the region $\Omega_p$.

Note that if a region $\Omega_p$ is dichotomized into two regions $\Omega_{p_0}$ and $\Omega_{p_1}$, then $Z(\Omega_p) \geq Z(\Omega_{p_0}) + Z(\Omega_{p_1})$. Also $Z(S) = \sum_{p=1}^{P}Z(\Omega_p)$ defines the total overall impurity of the space $S$.

We next proceed to discuss the process of hyperplane generation, which is a greedy approach. In this process we choose the most impure region $\Omega^{*}$ which is obtained as 

\begin{align}\label{eq::mostImpurePartitions}
\Omega^{*} = \arg\max_{\Omega_p \in \mathcal{F}_1} Z(\Omega_p),\:\text{where}
\end{align}
\begin{align}\label{eq::impurePartitions}
\mathcal{F}_1 = \{\Omega_p: \Omega_p\in \mathcal{F}, Z(\Omega_p)> 0 \text{ and } \exists j \text{ such that } ((m_{j}^{(p)}\neq M_{j}^{(p)})\}
\end{align}

consists of only impure regions that can be divided.

Let region $\Omega^{*}$ be divided into regions $\Omega^{*}_0$ and $\Omega^{*}_1$, where

\begin{align}\label{eq::partition1}
\Omega^{*}_0 = \{(x^{(i)}, y_i): \lambda^{*}(x^{(i)}) = 0\: \forall (x^{(i)}, y_i) \in \Omega^{*}\},
\end{align}

and

\begin{align}\label{eq::partition2}
\Omega^{*}_1 = \{(x^{(i)}, y_i): \lambda^{*}(x^{(i)}) = 1\: \forall (x^{(i)}, y_i) \in \Omega^{*}\}.
\end{align}

In (\ref{eq::partition1}) and (\ref{eq::partition2}), the mapping $\lambda^{*}$ is generated as for $\lambda_p$ defined at (\ref{eq::bitAssignment}). The mapping $\lambda_p$ is defined for all $\Omega_p$, and $\Omega^{*}$ is one of the $\Omega_p$'s from the family of $\mathcal{F}_1$.

The effect of the hyperplane corresponding to $\lambda^{*}$ is extended to other impure regions as well. For the region $\Omega_p \in \mathcal{F}_1 \setminus {\Omega^{*}}$, we define

\begin{align}\label{eq::remPartition1}
\Omega^{*}_{p_0} = \{(x^{(i)}, y_i): \lambda^{*}(x^{(i)}) = 0,\: (x^{(i)}, y_i) \in \Omega_p\},
\end{align}

and

\begin{align}\label{eq::remPartition2}
\Omega^{*}_{p_1} = \{(x^{(i)}, y_i): \lambda^{*}(x^{(i)}) = 1,\: (x^{(i)}, y_i) \in \Omega_p\},
\end{align}
so that $\Omega_p = \Omega^{*}_{p_0} \cup \Omega^{*}_{p_1}$ for $\Omega_p\in \mathcal{F}_1$ but $\Omega_p \neq \Omega^{*}$.

Next, $K$ different hyperplanes are generated via the procedure described in  (\ref{eq::minEquation})-(\ref{eq::biasP}) for the given region $\Omega^{*}$ as chosen from (\ref{eq::mostImpurePartitions}). These are denoted by $\langle w^{(k)},x\rangle + b^{(k)} = 0$, $k=1,2,..,K$. For each of these hyperplanes, the steps proposed in  (\ref{eq::partition1})-(\ref{eq::partition2}), and (\ref{eq::remPartition1})-(\ref{eq::remPartition2}), are performed, and $Z^{(k)}(S)$ is computed for $k=1,2,..,K$. Here $Z^{(k)}(S)$ is the notation used for $Z(S)$ with respect to the $k$th hyperplane $\langle w^{(k)},x\rangle + b^{(k)} = 0$, $k=1,2,..,K$. Let

\begin{align}\label{eq::minHyperplaneImpurity}
    Z^{(l)}(S) = \min_{k=1,2,..,K}(Z^{(k)}(S)).
\end{align}

We choose the hyperplane $\langle w^{(l)},x\rangle + b^{(l)} = 0$ and any tie in (\ref{eq::minHyperplaneImpurity}) is broken arbitrarily.

We subsequently update the family of impure regions $\mathcal{F}_1$ to take into account new nonempty impure regions. This gives a new updated family of impure regions. 

The process is repeated until no impure region is left to be further dichotomized.

Once the above process is completed, all pure regions are collected in the family $\Tilde{\mathcal{F}}$. Thus

\begin{align}\label{eq::allPureFilledPartitions}
\Tilde{\mathcal{F}} = \{\Omega_p :\Omega_p \in \mathcal{F}, Z(\Omega_p) = 0 \text{ or } m_{j}^{(p)}=M_{j}^{(p)} \forall j \in \{1,..,n\} \}.
\end{align}

Every pure region $\Omega_p$ in the family $\Tilde{\mathcal{F}}$ is assigned a code that is shared by every sample in the region. Here we assume that all regions have been placed in an arbitrary but fixed order $\Bar{\mathcal{F}}=(\Tilde{\mathcal{F}})$, then for any sample $(x^{(i)}, y_{i})\in S$, its $code_{x^{(i)}}\in\{0, 1\}^{r}, r \in \mathbb{N}$ is assigned as

\begin{align}\label{eq::codeAssiginment}
code_{x^{(i)}} = (\lambda^{p}(x^{(i)}):\forall \Omega_p \in \Bar{\mathcal{F}}),
\end{align}

where $r$ is the total number of hyperplanes.

The proportion of samples from different classes in resultant regions yields their probability. For a given test sample, these probabilities are combined across all instances, and it is associated with the class having the highest probability. Let us assume $f$ that maps every pure region (represented by its unique code) to the posterior probabilities of finding a class $c \in Y$ in the given region. In other words, let $f:\{0,1\}^r \times Y \to \mathbb{R}$, then

\begin{align}\label{eq::probAssignment1}
f{(code_{x^{(i)}}, y_{i})} = \frac{\hat{f}{(code_{x^{(i)}}, y_{i})} \times IF_{y_i}}{\sum_{c=1}^{C}IF_c \times \hat{f}{(code_{x^{(i)}}, c)}},
\end{align}

where

\begin{align}\label{eq::probAssignment}
\hat{f}{(code_{x^{(i)}}, y_{i})} =  \frac{|\{y_{j}:(y_{j}=y_{i})\land(code_{x^{(j)}}=code_{x^{(i)}}) \}\:\forall j \in \{1,..,N\}|}{|\{y_{j}:code_{x^{(j)}}=code_{x^{(i)}}\}\:\forall j \in \{1,..,N\}|},
\end{align}

and $IF_c$ denote the weight associated with a class $c$ such that abundant classes have smaller weights, and vice-versa.

\begin{align}
IF_c = \frac{N}{|\{y_{j}:y_{j}=c\}\:\forall j \in \{1,..,N\}|}\: \forall c \in \{1,...,C\}.
\end{align}

Let us define $h_t$ such that $h_t:X \times Y \to \mathbb{R}$, $\forall t \in \{1,...,{T}\}$. Further, we define $h_t$ as follows, that maps every pure region to its posterior probabilities.

\begin{align}\label{eq::totalHypothesis}
h_t(x^{(i)}, y_i) = f(code_{x^{(i)}}, y_i)\: \forall (x^{(i)}, y_i) \in X \times Y
\end{align}

The above steps outline the construction of one high variance classifier instance. It is well established in the literature, that an ensemble of such high variance instances, in general, tends to yield better generalization on test samples~\cite{schapire1998boosting}. Our proposed method GRAF creates several such high variance instances.

Next, we define $h$ such that it maps a sample to a class. This is done by using a consensus for prediction, that can be reached by computing the joint probability of predictions returned by each high variance classifier. We therefore define $h:X\to Y$ given by

\begin{align}\label{eq::prediction}
h(x^{(i)}) = \arg\max_{y_i \in Y}\sum_{t=1}^{T}\log_2\left(1+h_t(x^{(i)}, y_i)\right).
\end{align}

It should be noted, that when all regions for sample $x^{(i)}$ contain only one class $c$, then $h_t(x^{(i)},c)$ is 1 for $c$ and 0 for remaining classes. Hence, $h(x^{(i)})$ is equivalent to a voting classifier.

Given an ensemble of instances $h_1$, $h_2$,...,$h_T$, GRAF optimizes the margin function as follows

\begin{align}\label{eq::treeMargin}
mg(x^{(i)}, y_i) = \mathbbm{1}{\left(h(x^{(i)}) = y_i\right)} - \max_{y_j \in Y \setminus{y_i}}\mathbbm{1}{\left(h(x^{(i)})=y_j\right)}.
\end{align}

Hence, the margin over the complete set of samples $X \times Y$ is defined as

\begin{align}\label{eq::treeExpectedMargin}
mg = \textbf{E}_{X, Y}mg(x^{(i)}, y_i).
\end{align}

\section{Implementation details}\label{section::implementation}

\begin{figure}[!ht]
    \centering
    \includegraphics[width=\linewidth,height=6cm,keepaspectratio]{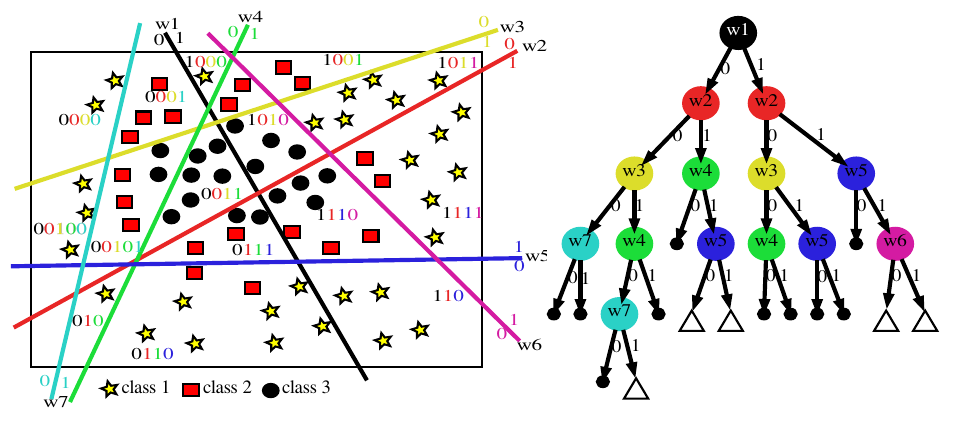}
    \caption{The division of space in GRAF is represented by a tree. A region containing a subset of samples is defined by its unique combination of hyperplanes. However, these hyperplanes may affect the formation of other regions. The process terminates once space is maximally divided such that the impurity in any region cannot be reduced any further. Every resultant region corresponds to a leaf node in the tree, represented by a dot in the figure. (A triangle denotes an impure region which may be dichotomized further.)   }
    \label{spacePartitioning}
\end{figure}

Guided random forest (GRAF) creates an ensemble classifier by repeatedly dichotomizing the input space. In order to build one classifier instance from a given set $S$ of samples, a subset of $M$ features is uniformly sampled from the given set of features $n$. Samples are then projected into this M-dimensional sub-space, denoted by $X_M$. To facilitate efficient implementation, the additive construction of an instance is represented as a tree from the beginning. The tree is represented by its collection of regions (Figure~\ref{spacePartitioning}). At the $0$-th height, $\Omega_{root}$ consist of all samples, $(x^{'})^{(i)} \in X_M$ and hence, the hyperplane $w^{\textit{(height)}}$ and $bias^{\textit{(height)}}$ is generated by considering all samples. At every height, the most impure region $\Omega^{*}$ (whole space at root), affects the generation of $w^{\textit{(height)}}$ and $bias^{\textit{(height)}}$. For $\Omega^{*}$, $K$ such hyperplanes are generated, and the effect of these hyperplanes is extended to other impure regions as well. The hyperplane whose inclusion yields the lowest overall space impurity $Z(S)$ is selected. Empty, pure and impure regions may exist at each given height. The number of these regions is given by $\sum_{i=0}^{M}\binom{height}{i}$ (for $height < i$, $\binom{height}{i} = 0$), i.e., it is a polynomial in $height$ of the order of $M$ ($\mathcal{O}(height^{M})$). Thus, the number of filled (pure and impure) regions is \\$\mathcal{O}(\min(N, \sum_{i=0}^{M}\binom{height}{i}))$. For further processing, only impure regions need to be considered. Hence, $\mathcal{F}_1$ consists of only impure regions. The most impure region $\Omega^{*} \in \mathcal{F}_1$ defines the distribution of the next random weight vector $w^{(\textit{height})}$ to be included at next height. Even though $w^{(\textit{height})}$ almost surely dichotomizes the region $\Omega^{*}$, it may or may not dichotomize other remaining regions in  $\mathcal{F}_1$. To avoid empty regions from being created, bit assignment is skipped for the non-dichotomized region at a given height. Hence, the resultant $code(j)$ for sample $x^{(j)}$ in region $\Omega_j$, formed by the concatenation of bits is of variable length. Once all impure regions have been fixed, leaf nodes represent the posterior probabilities of a class. The above procedure is repeated for the construction of other trees, with a different random sub-space of features of length $M$. Algorithm~\ref{algo::grafalgo} represents this process in a systematic manner. 

\subsection{Heuristic for region search}
A naive implementation of scanning-regions part of the algorithm will require scanning all the impure regions, which would incur an excessive overhead. GRAF employs a heuristic to limit the number of impure regions to be scanned.

The radius of influence (ROI) of a region $\Omega_{p}$ is defined as

\begin{align}\label{eq::roi}
    ROI_{p} = \max\Bigg(\sqrt{\sum_{j=1}^{j=n}(m^{(p)}_{j} - \mu^{(p)}_{j})^{2}}, \sqrt{\sum_{j=1}^{j=n}(M^{(p)}_{j} - \mu^{(p)}_{j})^{2}}\Bigg)
\end{align}

\begin{figure}[!ht]
    \centering
    \includegraphics[width=\linewidth,height=6cm,keepaspectratio]{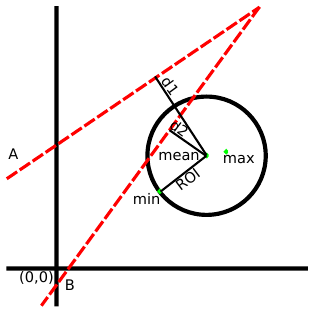}
    \caption{The perpendicular distance of mean point from plane A (d1) is greater than radius of influence (ROI). Hence, Plane A does not dichotomize the region. Perpendicular distance of the mean point from plane B (d2) is less than ROI. Hence, plane B may dichotomize the region. If the perpendicular distance is equal to ROI, it is considered as not dichotomized.}
    \label{fig::roi}
\end{figure}

A region is scanned for a split if the perpendicular distance (referred as $pdist$ in Algorithm~\ref{algo::grafalgo}) of the hyperplane to the mean (\ref{eq::meanEquation}) is less than ROI (\ref{eq::roi}). In Figure~\ref{fig::roi}, min corner of the region is farther away from the mean, and hence ROI is defined as the distance between these two points. Two hyperplanes A and B are shown, where the perpendicular distance of A from mean (d1) is greater than the ROI, and hence this plane is guaranteed not to split the regions. Therefore,  while scanning for hyperplane A, this region will be skipped. When the perpendicular distance of B from the mean (d2) is less than the ROI, hyperplane B may or may not split the region. Hence, the region will be scanned for hyperplane B.

\begin{algorithm}[!ht]
    \caption{GRAF algoritm}\label{algo::grafalgo}
    \begin{algorithmic}
        \State \textbf{Input:} Dataset $X\times Y$ containing $N$ samples of $n$ features
        \State \hspace{10mm} $T$ - total number of trees
        \State \hspace{10mm} $M$ - feature subspace size ($\leq n$)
        \State \hspace{10mm} $K$ - trials to search the most suitable hyperplane
        \State \textbf{for} $t=1$ to $T$ \textbf{do}
        \State \hspace{3mm} choose $M$-dimensional feature subspace $X_{M}$
        \State \hspace{3mm} $height \gets 0$
        \State \hspace{3mm} Create $\Omega_{root}$, a region of whole data $X_{M}$
        \State \hspace{3mm} $\Omega_{root}.lc \gets \varnothing$, $\Omega_{root}.rc \gets \varnothing$, $\Omega_{root}.bit \gets \varnothing$ 
        \State \hspace{3mm} $\Omega_{root}.p \gets \varnothing$, $\Omega_{root}.h \gets height$
        \State \hspace{3mm} $\Omega_{root}.roi \gets ROI_{root}$ (\ref{eq::roi})
        \State \hspace{3mm} $\mathcal{F}_{1} \gets \{ \Omega_{root} \}$, $\Omega^{*} \gets \Omega_{root}$
        \State \hspace{3mm} \textbf{while} $|\mathcal{F}_{1}| > 0$ \textbf{do}
        \State \hspace{6mm} $W, b \gets$  generate $K$ hyperplanes for $\Omega^{*}$ (\ref{eq::weightGeneration}, \ref{eq::biasP})
        \State \hspace{6mm} \textbf{for} $k \in \{1,..,K\}$ \textbf{do}
        \State \hspace{9mm} $\mathcal{F}^{k} \gets \varnothing$
        \State \hspace{9mm} split $\Omega^{*}$ into $(\Omega^{*}_{0})^{k}$ and $(\Omega^{*}_{1})^{k}$ (\ref{eq::partition1}, \ref{eq::partition2})
        \State \hspace{9mm} $(\Omega^{*}_{0})^{k}.bit \gets 0$, $(\Omega^{*}_{1})^{k}.bit \gets 1$
        \State \hspace{9mm} $(\Omega^{*}_{0})^{k}.p \gets \Omega^{*}$, $(\Omega^{*}_{1})^{k}.p \gets \Omega^{*}$
        \State \hspace{9mm} $(\Omega^{*}_{0})^{k}.h, (\Omega^{*}_{1})^{k}.h \gets height + 1$
        \State \hspace{9mm} $\mathcal{F}^{k} \gets \mathcal{F}^{k}\cup\{(\Omega^{*}_{0})^{k}, (\Omega^{*}_{1})^{k}\}$
        \State \hspace{9mm} \textbf{for} $\Omega_{p} \in \mathcal{F}_{1} \setminus \{\Omega^{*}\}$
        \State \hspace{12mm} \textbf{if} $\Omega_{p}.roi > pdist(W^{k}, \mu^{p})$ \textbf{then}
        \State \hspace{15mm} split $\Omega_{p}$ into $(\Omega^{*}_{p0})^{k}$ and $(\Omega^{*}_{p1})^{k}$ (\ref{eq::remPartition1}, \ref{eq::remPartition2})
        \State \hspace{15mm} \textbf{if} $|\Omega^{*}_{p0}| > 0$ \& $|\Omega^{*}_{p1}| > 0$  \textbf{then}
        \State \hspace{18mm} $(\Omega^{*}_{p0})^{k}.p \gets \Omega_{p}$, $(\Omega^{*}_{p1})^{k}.p \gets \Omega_{p}$
        \State \hspace{18mm} $(\Omega^{*}_{p0})^{k}.bit \gets 0$, $(\Omega^{*}_{p1})^{k}.bit \gets 1$
        \State \hspace{18mm} $(\Omega^{*}_{p0})^{k}.h, (\Omega^{*}_{p1})^{k}.h \gets height + 1$
        \State \hspace{18mm} $\mathcal{F}^{k} \gets \mathcal{F}^{k}\cup\{(\Omega^{*}_{p0})^{k}, (\Omega^{*}_{p1})^{k}\}$
        \State \hspace{15mm} \textbf{else}
        \State \hspace{18mm} $\mathcal{F}^{k} \gets \mathcal{F}^{k} \cup \{\Omega_{p}\}$
        \State \hspace{12mm} \textbf{else}
        \State \hspace{15mm} $\mathcal{F}^{k} \gets \mathcal{F}^{k} \cup \{\Omega_{p}\}$
        \State \hspace{9mm} compute impurity of resultant partition of $S$ as $Z^{k}(S) = \sum_{\Omega_{p}\in\mathcal{F}^{k}}Z(\Omega_{p})$
        \State \hspace{6mm} $bestK \gets \arg \min_{k \in \{1,..,K\}}Z^{k}(S)$
        \State \hspace{6mm} $w^{(height)} \gets W^{(bestK)}$
        \State \hspace{6mm} $bias^{(height)} \gets b^{(bestK)}$
        \State \hspace{6mm} $\mathcal{F}_{1} \gets \mathcal{F}^{(bestK)}$
        \State \hspace{6mm} \textbf{for} $\Omega_{p} \in \mathcal{F}_{1}$ \textbf{do}
        \State \hspace{9mm} \textbf{if} $\Omega_{p}.bit = 0$ \textbf{then} $\Omega_{p}.p.lc \gets \Omega_{p}$
        \State \hspace{9mm} \textbf{if} $\Omega_{p}.bit = 1$ \textbf{then} $\Omega_{p}.p.rc \gets \Omega_{p}$
        \State \hspace{6mm} $height \gets height + 1$
    \end{algorithmic}
\end{algorithm}

\subsection{CPU vs GPU implementation}
For each impure region in $\mathcal{F}_1$, the division of region (\ref{eq::partition1}-\ref{eq::partition2}) requires a multiplication of two matrices of size $n_p \times M$ and $M \times K$. Matrix multiplication is computationally intensive, requiring $\mathcal{O}(n_p\times M\times K)$ CPU operations. Graphical Processing Units (GPUs) can significantly reduce matrix multiplication time via parallel computation.

GRAF's GPU implementation differs slighly from the CPU one. To avoid massive data transfer between the host's RAM and the GPU co-processor, all training samples ($N \times M$) are stored in the GPU's RAM before initiating the training process. Upon selection of $\Omega^{*}$, the generated weight matrix of size $M \times K$ and the bias vector of length $K$ is sent to the GPU, and a region assignment matrix of size $N \times K$ is retrieved. All impure regions from $\mathcal{F}_1$ are then scanned, to find the overall reduction in impurity ($Z(S)$) to select the best hyperplane.

\subsection{Time Complexity}\label{sec::timeComplexity}

To analyse the worst case time complexity, assume a dataset where the neighborhood of each sample consists of examples from different classes. Further, assume that full trees are grown, and that there are $N$ samples with $M$ dimensions. In this case, all leaf nodes will contain only one sample. Hence, there will be $N$ leaf nodes in the tree.

\subsubsection{Training time complexity of a tree}

Let us first assume that balanced trees are grown. In this case, the maximum number of impure regions at any time would be $N/2$. In the worst case, each hyperplane will only divide the region for which it was generated. The scanning of the region will take $\mathcal{O}(\sum_{i=1}^{i=(N/2)-1}(K\times N))$ time, until the maximum number of impure regions is created. Subsequently generated hyperplanes will "purify" at least one region. This will take $\mathcal{O}(\sum_{i=1}^{i=N/2}(K\times N))$ time. Hence, the total time spent in scanning will be $\mathcal{O}(K\times(N^2-N))\equiv\mathcal{O}(K\times N^2)$. Therefore, the total number of generated weights will be $N-1$ (total number of non leaf nodes). The total time spent in matrix multiplication will be $\mathcal{O}((N\times M\times K+K)\times(N-1))$. Hence, total train time complexity $\mathcal{O}((N\times M\times K+K)\times(N-1) + K\times N^2)$.

In another scenario, assume that extremely skewed trees are generated. The maximum number of impure regions at any time will be $1$. In this case, the total number of generated weights will be $N-1$, and total train time complexity is given as $\mathcal{O}((N\times M\times K+K)\times(N-1)+K)$.

The above mentioned cases represent extreme scenarios. In practice, the training time complexity of GRAF will lie somewhere in-between. Let the the total number of generated weights be denoted by $TW$. Since weights are shared between regions, the value of $TW$ will be much smaller than $N-1$, and matrix multiplication time will reduce to $\mathcal{O}((N\times M\times K+K)\times TW)$. Similarly, the maximum number of impure regions at any instance is much smaller than $N/2$, since samples from similar classes tend to cluster. This reduces the total number of leaf nodes, which in turn reduces the maximum number of non leaf nodes needed to be searched at any instance. The time required to scan impure regions can be reduced further by ROI heuristic. With the ROI heuristic, only a fraction of impure regions need to be scanned to compute the quality of a hyperplane. However, this value is still upper bounded by $\mathcal{O}(K\times N^2)$. 

Hence, the worst case train time complexity of GRAF for a CPU implementation is $\mathcal{O}((N\times M\times K+K)\times TW + K\times N^2)$. Since matrix multiplication can be parallelized with GPUs, the time complexity for a GPU implementation is given by $\mathcal{O}(C_1 TW + C_2 + K\times N^2)$, where $C_1$ and $C_2$ are overheads for weight transfer, and data transfer, respectively. 

\subsubsection{Testing time complexity of a tree}
The worst case test time complexity of GRAF is defined as the total time taken to reach a leaf node. For a given test sample, it is equal to $\mathcal{O}(max\_tree\_height\times M)$ for a CPU implementation. For a GPU implementation, it is $\mathcal{O}($ $max\_tree\_height + C_1)$, where $C_1$ is data transfer overhead.

\subsection{Model Size}\label{sec::modelsize}
The model size of GRAF corresponds to the amount of information needed to make predictions. Since GRAF uses a binary tree data structure, every internal/non-leaf ($TNL$) node will have exactly two child nodes. In addition, it also contains information about the index of weight to decide which path to traverse. Each leaf node ($TL$) also contains label information. Hence, the total model size (for a tree) of GRAF is given by $TW\times (M+1) + TNL \times 3 + TL$.

\subsection{Space Complexity}
The scenario as described in Section~\ref{sec::timeComplexity} is followed to discuss the space complexity of GRAF. In addition to the space required to store a dataset, GRAF requires $\mathcal{O}(N)$ space to store temporary regions spawned in every trial. To perform $K$ trials, the total space requirement is $\mathcal{O}(K\times N)$. GRAF also needs to store the tree in memory. As disscussed in section~\ref{sec::modelsize},  the total space required to store a tree is $TW\times (M+1) + TNL \times 3 + TL$, and hence, the total space complexity of GRAF is $\mathcal{O}(K\times N + TW\times (M+1) + TNL \times 3 + TL)$

\section{Relationship of GRAF with boosting}\label{sec:graf_adaboost}
As shown in Algorithm~\ref{algo::bsummary}, the construction of a high variance instance of a classifier can be abstracted as a boosting algorithm~\cite{freund1997decision}. Assuming that the weight of each sample is initially $1$, a random hyperplane is generated (\ref{eq::bitAssignment}). This generated hyperplane divides the region into two parts. Sample weights are updated to focus on the region under consideration, based on their impurity (\ref{eq::labeldef}). All the samples in that region are assigned a weight of $1$, while remaining samples are assigned a weight of $0$. A new random hyperplane is generated (\ref{eq::weightGeneration}) based on the weight distribution of samples. However, this new plane is extended to other regions as well. The combination of all these planes (hypotheses) increases confidence, and hence, eventually creates a strong learner.

\begin{algorithm*}[!ht]
    \caption{High variance instance of GRAF as boosting}\label{algo::bsummary}
    \begin{algorithmic}
        \State \textbf{Input:}$(x^{(1)}, y_1)$,..,$(x^{(N)}, y_N)$; $x^{(i)} \in X$, $y_i \in \{1,..,C\}$, $C$ denotes the total unique classes and $N$ denotes the total training samples.
        \State \hspace{10mm}$Z:\mathcal{F}\to\mathbb{R}$ where $\Omega \in \mathcal{F}$ constitutes a set of points with same code.
        \State \hspace{10mm}$Y=\{1,..,C\}$
        \State \textbf{Initialize:} $P(i) \gets 1\: \forall i \in \{1,..,N\}$
        \State \hspace{15mm}$\textit{code}(i) \gets \varnothing\: \forall i \in \{1,..,N\}$
        \State \textbf{until} $\sum_{i=1}^{i=N} P(i) = 0$ \textbf{do}
        \State \hspace{8mm}Choose a random hypothesis using $P(i)$, such that $\lambda: X \to \{0,1\}$
        \State \hspace{8mm}$\textit{code}(i) \gets \textit{code}(i) \cup \{\lambda(x^{(i)})\}\: \forall i \in \{1,..,N\}$
        \State \hspace{8mm}Let $\Omega{_i} \gets \{(x^{(j)}, y_j): code(j) = code(i)\:\forall j \in \{1,..,N\}\}\:\forall i \in \{1,..,N\}$
        \State \hspace{8mm}$\omega \gets \arg\max_{i \in \{1,..,N\}} Z(\Omega_{i})$
        \State \hspace{8mm}Update $P(i) \gets \mathbbm{1}{\left(\Omega_{i}=\Omega_{\omega}\right)}\: \forall i \in \{1,..,N\}$
    \end{algorithmic}
\end{algorithm*}

\section{Simulation Study}\label{section::simulation}

A simulation study was designed to discuss the design aspects of GRAF, such as oblique hyperplanes for dichotomization, and extension of the hyperplane. It is known that axis-aligned decision trees do not generalize well for tasks with high concept variation~\cite{perez1996learning,bengio2010decision}. To emulate a high concept variation task, samples were generated near the vertices of a $n$ dimensional hypercube as per Algorithm~\ref{algo::simulationData}. For a binary classification task, the parity function was considered. A label 1 is assigned to a sample if it is generated near a vertex having odd number of 1's, and a label 0 otherwise. For a multi-class classification task, the label is assigned as the total number of 1's in the neighbouring vertex.

\begin{algorithm}[!ht]
    \caption{Simulation Data}\label{algo::simulationData}
    \begin{algorithmic}
        \State \textbf{Input:} $n$ dimension of hypercube.
        \State \textbf{Initialize:} 
        \State \hspace{4mm} $sample\_per\_vertex \gets [3, 4, 5]$
        \State \hspace{4mm} $all\_coords \gets$ all vertices of n dimensional hypercube
        \State \hspace{4mm} $mean\_0$, $mean\_1$, $stdev\_0$ and $stdev\_1$ of size $n$
        \State \textbf{Output:} $generated\_data \gets []$ 
        \State \textbf{Run:}
        \State \textbf{for} $i \in \{1,..,n\}$ \textbf{do}
        \State \hspace{4mm}$mean\_0_i$, $stdev\_0_i \sim \mathcal{U}[-0.5, 0.5)$
        \State \hspace{4mm}$mean\_1_i$, $stdev\_1_i \sim \mathcal{U}[0.5, 1.5)$
        \State \textbf{for} $coord \in all\_coords$ \textbf{do}
        \State \hspace{4mm} $c \gets $ \text{select one number randomly from } $sample\_per\_vertex$
        \State \hspace{4mm} \textbf{for} $j \in \{1,..,c\}$ \textbf{do}
        \State \hspace{8mm} $gen\_sample \gets $ array of size $n$
        \State \hspace{8mm} $ct \gets 0$
        \State \hspace{8mm} \textbf{for} $bit \in coord$ \textbf{do}
        \State \hspace{12mm} \textbf{if} $bit = 0$ \textbf{then}
        \State \hspace{16mm} $gen\_sample_{ct}$ $\sim$ $\mathcal{N}$($mean\_0_{ct}$, $stdev\_0_{ct}$) until $-0.5 <$ $gen\_sample_{ct}$ $< 0.5$
        \State \hspace{12mm} \textbf{if} $bit = 1$ \textbf{then}
        \State \hspace{16mm} $gen\_sample_{ct}$ $\sim$ $\mathcal{N}$($mean\_1_{ct}$, $stdev\_1_{ct}$) until $0.5 <$ $gen\_sample_{ct}$ $< 1.5$
        \State \hspace{12mm} $ct \gets ct + 1$
        \State \hspace{8mm} $generated\_data.append(gen\_samples)$
    \end{algorithmic}
\end{algorithm}

The number of features ($n$) is varied from 3 to 15 (since very few samples can be generated when only 2 features are used). In effect, the total number of samples vary from $\sim25$ - $\sim115,000$ (Table~\ref{tab:simulation_data}). For a multiclass example with $n$ features, $n+1$ classes are possible. For a given configuration (binary or multiclass) with $n$ features, 10 different datasets were generated. For every dataset, the train-test split consisted of 70-30\% of the total samples. 

\begin{table}[!ht]
    \centering
    \scalebox{0.9}{
    \begin{tabular}{||c|c|c|c|c||}
        \hline
        Features & Classes & Train samples & Test Samples & PC(v=0.9)\\
        \hline\hline
        3 & 2,4 & 18.1$\pm$0.700 & 8.6$\pm$0.489&3,1.7\\
        4 & 2,5 & 38.9$\pm$1.044 & 17.4$\pm$0.663&4,1.5\\
        5 & 2,6 & 77.8$\pm$1.887  & 34.3$\pm$0.900&5,2.1 \\
        6 & 2,7 & 155.6$\pm$1.685 & 67.3$\pm$0.900&6,2.2 \\
        7 & 2,8 & 312.9$\pm$3.477 & 134.6$\pm$1.497&7,2.6 \\
        8 & 2,9 & 626.5$\pm$5.463 & 269.3$\pm$2.452&8,2.7 \\
        9 & 2,10 & 1256.5$\pm$9.729 & 539.1$\pm$4.346&8,2.9 \\
        10 & 2,11 & 2515.4$\pm$10.312 & 1078.7$\pm$4.647&9,3.3 \\
        11 & 2,12 & 5024.7$\pm$15.408 & 2154.1$\pm$6.730&10,3.3 \\
        12 & 2,13 & 10032.9$\pm$10.540 & 4300.6$\pm$4.652&11,3.6 \\
        13 & 2,14 & 20072.4$\pm$36.546 & 8603.1$\pm$15.776&12,3.8 \\
        14 & 2,15 & 40129.0$\pm$41.613 & 17198.8$\pm$17.713&13,4 \\
        15 & 2,16 & 80302.7$\pm$68.444 & 34416.3$\pm$29.312&14,4.5 \\
        \hline\hline
    \end{tabular}
    }
    \caption{A simulation study to discuss the design aspects of GRAF. The number of features was varied from 3 to 15. For a given value of the feature, both binary and multiclass examples were generated. For every configuration, 10 different trials were performed to generate samples. The total number of samples vary from $\sim25$ - $\sim115,000$ across all trials. The train-test split consists of 70-30\% of the total samples. The total number of principal components which explain 90\% of the total variance in the dataset differs when it is projected on a random matrix.}
    \label{tab:simulation_data}
\end{table}

\begin{figure}[!ht]
    \centering
    \includegraphics[width=\linewidth,keepaspectratio]{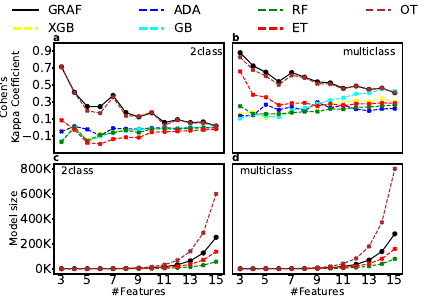}
    \caption{The performances of methods are compared on simulated binary and multiclass examples. The number of features varies from 3 to 15. For both binary and multiclass examples, GRAF has the highest values of Cohen's kappa coefficients, closely followed by Oblique Tree (OT). However, for similar performance measures, the overall model size of OT is much higher when compared with GRAF.}
    \label{fig::simulationIndividual}
\end{figure}

For comparison, $100$ trees were generated for every method, and the entire feature space was considered for every tree. For all experiments, $K$ (for GRAF) was equal to $M$ and $M=n$. For a given feature ($n$) and label information (binary or multi-class), the performance of a method was evaluated using Cohen's kappa coefficient for every trial, and averaged across all trials. For both binary and multiclass cases, the performance of GRAF supercedes others, closely followed by Oblique Tree (OT)~\cite{menze2011oblique} (Figure~\ref{fig::simulationIndividual}a-b). This is primarily because when concept variation is high, all features are independent and relevant. Thus, axis-aligned decision trees suffer because they consider only a single feature at a time to define a region. The performances of all others such as Adaboost (ADA)~\cite{freund1997decision}, Random Forest (RF)~\cite{breiman2001random}, XGBoost (XGB)~\cite{chen2016xgboost}, Gradient Boosting (GB)~\cite{friedman2001greedy}, and Extremely Randomized Trees (ET)~\cite{geurts2006extremely} are comparable to each other. The model size of ET, RF, GRAF, and OT has also been compared. For decision trees, the model size is mainly affected by factors such as the total number of internal/non-leaf nodes (TNL), the total number of leaf nodes (TL), and the total weights generated (TW). Non-leaf nodes contain threshold information, links to both child nodes, and the feature used for the split. The leaf nodes contain label information. For ET, RF, and OT, the total number of weights is equal to the total number of non-leaf nodes in the tree. The overall model size for ET and RF is $TNL\times 4+TL$. For OT, the weight vector $w$ lies in $\mathbb{R}^n$. Hence, the model size of OT is $TNL\times (n+1)+TNL\times2+TL$. However, for GRAF, since weights are shared between different regions, the total number of weights is much smaller than the total number of non-leaf nodes. GRAF's model size is therefore $TW\times (n+1)+TNL\times 3+TL$. GRAF's model size is significantly smaller than OT's, for comparable performance (Figure~\ref{fig::simulationIndividual}c-d).

\begin{figure}[!ht]
    \centering
    \includegraphics[width=\linewidth,keepaspectratio]{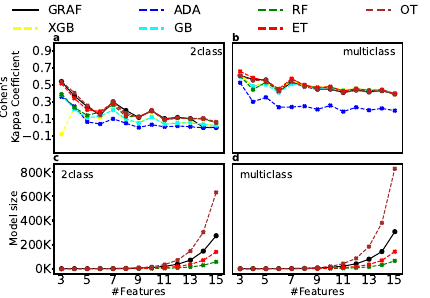}
    \caption{The performance of methods is compared when only a few features are relevant and independent. The performances of all methods are comparable. }
    \label{fig::simulationProjection}
\end{figure}

The essence of the previous simulation study was to establish the fact, that for a scenario where all features are independent and relevant, GRAF shows satisfactory performance along with a competitive model size. In addition to this, it is imperative to evaluate the performances of methods when all features are not necessarily independent. For this, the samples in the previous study are projected by using a random matrix. The resultant dataset has its overall variance explained with a few principal components (Table~\ref{tab:simulation_data}). For instance, when $15$ features are used to generate a simulated dataset, $14$ principal components are needed to explain 90\% of the total variance in the dataset. On the other hand, when the same dataset is projected by using a random matrix, less than $5$ principal components are adequate. For this scenario, similar experiments were performed. Almost all methods have comparable performances (Figure~\ref{fig::simulationProjection}) for this case. In other words, GRAF gives satisfactory performances in both scenarios. 

\begin{figure}[!ht]
    \centering
    \includegraphics[width=\linewidth,keepaspectratio]{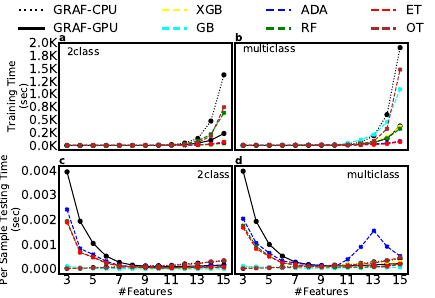}
    \caption{The training and testing time of different methods are compared on a simulated dataset. GRAF's GPU implementation significantly reduces the training time for both binary and multiclass examples. GRAF's testing time is comparable with other methods. }
    \label{fig::simulationTimeIndividual}
\end{figure}

\begin{figure}[!ht]
    \centering
    \includegraphics[width=\linewidth,keepaspectratio]{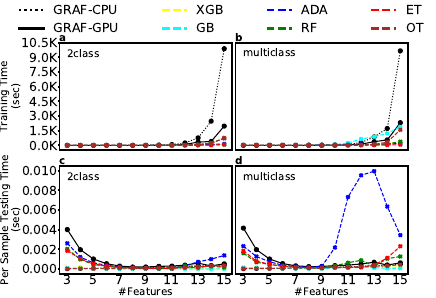}
    \caption{The training and testing times of different methods are compared on a simulated dataset projected by using a random matrix. The GPU implementation of GRAF significantly reduces its training time for both binary and multiclass examples. The testing time of GRAF is comparable with other methods. }
    \label{fig::simulationTimeProjection}
\end{figure}

The other important criterion to compare different methods is their run time complexity. As discussed in section~\ref{sec::timeComplexity}, GRAF's GPU train time (GRAF-GPU) is considerably lower than its CPU counterpart, because GRAF involves matrix multiplication. Hence, we compare the training and test time complexity of both implementations of GRAF with OT, ET, RF, GB, ADA and XGB on simulated dataset (Figure~\ref{fig::simulationTimeIndividual}) and simulated dataset after projection (Figure~\ref{fig::simulationTimeProjection}). As shown in Figure~\ref{fig::simulationTimeIndividual}a-b, the training time of GRAF-GPU is considerably smaller than OT and GB, and competitive with RF and XGB. GRAF-GPU's test time (Figure~\ref{fig::simulationTimeIndividual}c-d) is higher for smaller datasets, because the data transfer overhead overshadows the speed gain from parallelization, while being considerably smaller for larger datasets.  

In all the above experiments, the number of trials for GRAF is equal to the number of features in the dataset. It was also observed, that the performance of GRAF without trials is slightly lower when compared with its trial counterpart. However, the training time is significantly lower. For the cases where features are independent and informative, the training time of GRAF is as fast as ET.

Performance measures reported in this article are recorded on a workstation with 40 cores using Intel\textregistered Xeon\textregistered E7-4800 (Haswell-EX/Brickland Platform) CPUs with a clock speed of 1.9 GHz, 1024 GB DDR4-1866/2133 ECC RAM and Ubuntu 14.04.5 LTS operating system with 4.4.0-38-generic kernel. The time taken by each algorithm has been measured by running it on a single core. For computation on GPU, 12GB NVIDIA Tesla K80 GPU is used.

This simulation study explains that cases where features are independent and relevant, oblique partitions (GRAF, OT) fair well in comparison to axis-aligned (RF, ET) partitions (Figure~\ref{fig::simulationIndividual}a-b). However, in the cases where the intrinsic dimensionality of data is smaller in comparison to the number of features, all methods have comparable performance (Figure~\ref{fig::simulationProjection}a-b). These results are concordant with the previously observed results~\cite{menze2011oblique,zhang2017robust}. Between GRAF and OT, GRAF has a smaller model size. This is because in GRAF, hyperplanes are shared between multiple regions, while in OT, each hyperplane does local partitioning. Therefore, GRAF has fewer hyperlanes and hence, a smaller model size.  However, ET and RF have lower model size in comparison to GRAF (Figure~\ref{fig::simulationIndividual}c-d, \ref{fig::simulationProjection}c-d). In the first case, the training time of GRAF-GPU is lower in comparison to OT and RF (Figure~\ref{fig::simulationTimeIndividual}a,b) but in a later case, the training time of GRAF-GPU is the highest (Figure~\ref{fig::simulationTimeProjection}a,b). All methods have equivalent testing time (Figure~\ref{fig::simulationTimeIndividual}c,d, \ref{fig::simulationTimeProjection}c,d). Considering all these aspects, it may be concluded that for the first case, GRAF can be a choice of method for both binary and multiclass cases. 

\section{Results}\label{section::results}
\subsection{Bias-variance tradeoff}\label{section::bias-variance}
In order to understand the behavior of a classifier, it is imperative to study its bias-variance tradeoff. A classifier with a low bias has a higher probability of predicting the correct class than any other class, i.e., the predicted output is much closer to the true output. On the other hand, the classifier with low variance indicates that its performance does not deviate for a given test set across several different models. There are several methods to evaluate bias-variance tradeoff for 0-1 loss on classification learning~\cite{breiman1996bias,kohavi1996bias,domingos2000unified,james2003variance}. Of these, we use the definitions of Kohavi \& Wolpert~\cite{kohavi1996bias} for bias-variance decomposition (\ref{eq::vote}-\ref{eq::err}).

\begin{align}\label{eq::vote}
p^{(i)}_j=\frac{1}{R}\sum_{r=1}^{r=R}\mathbbm{1}{(\hat{y_{i}}=j}) 
\end{align}
\begin{align}\label{eq::bias}
bias^2 = \frac{1}{N_t}\Bigg(\sum_{i=1}^{i=N_t}\sum_{j=1}^{j=C}\big((\mathbbm{1}{(y_{i}=j)} - p^{(i)}_j)^{2} - \frac{p^{(i)}_j*(1-p^{(i)}_j)}{R-1}\big)\Bigg)
\end{align}
\begin{align}\label{eq::variance}
variance = 1 - \frac{1}{N_t}\sum_{i=1}^{i=N_t}\sum_{j=1}^{j=C}(p^{(i)}_j)^{2}
\end{align}
\begin{align}\label{eq::err}
err = \frac{1}{R}\sum_{r=1}^{r=R}\Big(1-\frac{1}{N_t}\sum_{i=1}^{i=N_t}\mathbbm{1}{(y_{i}=\hat{y_{i}})}\Big)
\end{align}

For the analysis of bias-variance tradeoff, $N/2$ samples were set aside as the test set. From the remaining dataset, $R$ overlapping training sets of the same size $N_m$ were created, and $R$ models were trained. For every model, the estimate $\hat{y_i}$ is obtained for every instance $i$ in the test set, whose size is denoted by $N_t$.

Two different studies were performed to evaluate the performance of GRAF in terms of bias and variance decomposition. First, the effect of different values of hyper-parameters (namely, number of trees and feature sub-space size) on the bias, variance, and the misclassification error rate was analyzed. Second, the trends of bias and variance were observed for increasing train set sizes and compared with different classifiers. To perform these analyses, 6 different binary and multi-class datasets with a different number of centroids from \{10, 20, 50\} were simulated with Weka~\cite{witten1999weka}\footnote{Commands to generate a dataset, and their description are available in section~\ref{sec:dataGeneration}}. Each dataset consisted of 10000 samples and 10 features (generated using RandomRBF class), while other parameters were set as default. To create the test set, $5000$ samples were randomly selected. For a given train dataset size ($200 \leq N_m \leq 2500$), 50 models were generated by repeatedly sampling without replacement, from the remaining dataset.

\begin{figure}[!ht]
    \centering
    \includegraphics[width=\linewidth,keepaspectratio]{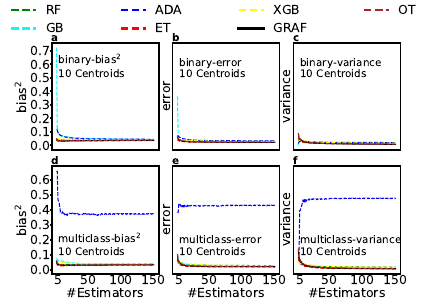}
    \caption{Bias-variance analysis with an increasing number of estimators (trees) in a classifier. For both binary (a-c) and multi-class (d-f) datasets with 10 centroids, the number of estimators is increased from $2$ to $150$, while fixing the number of dimensions to be sampled $(M = n/2)$. As the number of estimators is increased, bias, error, and variance rapidly saturate.}
    \label{increasingL}
\end{figure}

\begin{figure}[!ht]
    \centering
    \includegraphics[width=\linewidth,keepaspectratio]{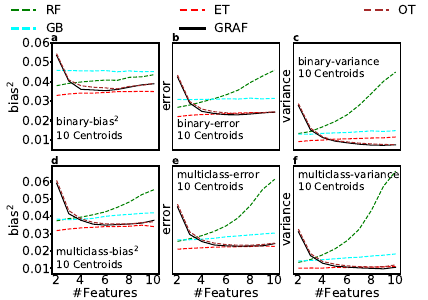}
    \caption{Bias-variance analysis with an increasing number of dimensions (features) selected from a given feature space in a classifier. For both binary (a-c) and multi-class (d-f) datasets with 10 centroids, $M$  is increased from $2$ to $10$, while fixing the number of estimators to be assembled $(L = 100)$. For GRAF, when the dimension of the sub-space is large enough to distinguish samples of different classes, bias and variance saturate and converge to their minimum. With increasing dimensionality of the sub-space, misclassification error continues to decrease and rapidly saturates to its minimum.}
    \label{increasingM}
\end{figure}

The effect of increasing the number of trees from 2 to 150 for 10 centroids is illustrated in Figure~\ref{increasingL} (Figures~\ref{increasingL20} and \ref{increasingL50} for 20 and 50 centeroids, respectively). For intermediate values of tree numbers, bias-variance curves saturate to their minima, and hence, the average misclassification converges to its minimum. It implies that higher accuracies can be achieved well before all trees are used~\cite{ho1998random}.  Figure~\ref{increasingM} highlights the effect of increasing the number of randomly selected dimensions/features for 10 centeroids (Figures~\ref{increasingM20} and \ref{increasingM50} for 20 and 50 centeroids, respectively). This figure shows that a subset of features, in general, may be enough to generate the desired results. However, the selected sub-space must be large enough to distinguish the samples in this sub-space. For these experiments $N_m$ was set to $2500$.

\begin{figure}[!ht]
    \centering
    \includegraphics[width=\linewidth,keepaspectratio]{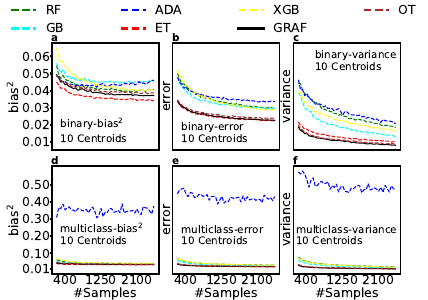}
    \caption{Bias-variance analysis with an increasing samples in a training set. For both binary (a-c) and multi-class (d-f) datasets with 10 centroids, the number of samples is increased from $200$ to $2500$, while fixing the number of dimensions to be sampled $(M = n/2)$ and the number of estimators as $L=100$. As the cardinality of the training set is increased, bias-variance continues to decrease, and the misclassification error continues to decrease and may saturate to its minimum.}
    \label{comparison50}
\end{figure}

In a different study, the influence of an increasing number of training samples ($200 \leq N_m \leq 2500$) is illustrated in Figure~\ref{comparison50} for a dataset with 10 centroids (Figures~\ref{comparison20} and \ref{comparison50} are for 20 and 50 centroids, respectively). Bias and variance decrease with an increase in the size of the training set. In general, GRAF was found to have the least variance, and the lowest or comparable misclassification errors on test samples, when compared with other methods (default values of hyper-parameters are used, $L=100$ and $M=5$).

\subsection{Performance comparison on UCI datasets}
The performance of GRAF has been evaluated on 115 UCI datasets~\cite{Dua:2017} and compared against random forest (RF)~\cite{breiman2001random}, gradient boosting (GB)~\cite{friedman2001greedy}, adaboost (ADA)~\cite{freund1997decision}, extremely randomized trees (ET)~\cite{geurts2006extremely}, xgboost (XGB)~\cite{chen2016xgboost}, and oblique tree (OT)~\cite{menze2011oblique}. Statistics of all 115 datasets are available in Table~\ref{dataStat}. The total number of samples across all datasets varies from 24 to $\sim 130$k. The count of features across datasets varies from 3 to 262. For comparison, we used the strategy as defined in Fernandez-Delgado \textit{et al.}~\cite{fernandez2014we}\footnote{Fernandez-Delgado et al. concluded that random forest is the best performing algorithm after comparing 179 classifiers. These results may be found at \url{http://persoal.citius.usc.es/manuel.fernandez.delgado/papers/jmlr/ data.tar.gz}}. They use four-fold cross-validation on the whole dataset to compute the performance. The training dataset contains 50\% of the total samples. 

The hyper-parameters are tuned using 5-fold cross-validation on the training dataset. For all methods, the number of estimators is tuned from $\{100, 200, 500, 1000, 2000\}$. For GRAF, RF, GB, ET, and OT, the number of dimensions to be selected ($M$) has been tuned from $\{\log_2(n), \sqrt{n}, n/2, n\}$, and the node is further split only if it has minimum samples, tuned between 2 and 5. For GRAF and OT, the number of trials (hyperplane search) $K$ is set to the value of $M$.

The average of the test set Cohen's kappa score across 4-folds of cross-validation has been tabulated in Table~\ref{kappaUCI}. For every dataset, the method with the highest score has been highlighted. On 33 datasets, GRAF outperforms all other methods. On 87, 66, 77, 71, 101, and 77 datasets, GRAF's performance is either better than or comparable with OT, ET, RF, GB, ADA, and XGB, respectively. 

As discussed in section~\ref{section::simulation}, oblique partitioning based trees have a better performance where features are independent and relevant in comparison to axis-aligned partitioning based trees. To reinforce this, we extend this analysis to UCI datasets as well. Table A11 contains the information about number of principal components (PC) required to explain the 50\%, 70\% and 90\% variance in columns PC(v=0.5), PC(v=0.7) and PC(v=0.9), respectively. GRAF has improved performance on datasets (PC(v=0.9)/total features) with a large number of components to explain the high variance, such as adult (12/14), balance-scale (4/4), bank (13/16), congressional-voting (11/16), mammographic (4/5), statlog-australian-credit (11/14), titanic (3/3), waveform (15/21), wine-quality-red (7/11), yeast (7/10), led-display (6/7), etc. when compared with ET and RF. On the other hand, GRAF has either poor or comparable performance on miniboone (2/50), musk-1 (23/66), musk-2 (26/166), statlog-landsat (4/36), plant-margin (25/64), plant-shape (2/64), plant-texture (20/64), etc.

\begin{figure}
    \centering
    \includegraphics[width=\linewidth,keepaspectratio]{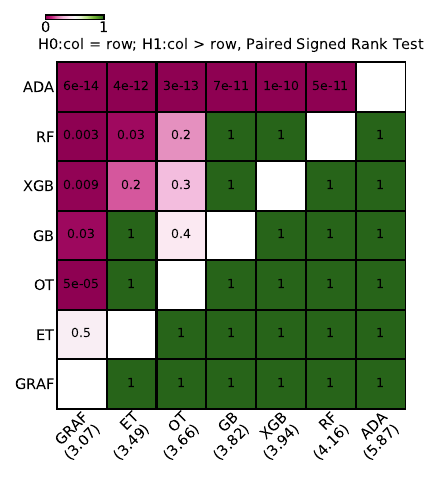}
    \caption{One-sided paired Wilcoxon signed-rank test on Cohen's kappa score. Each method is paired with every other method, and p-value was computed for the null hypothesis '$\text{left method} = \text{right method}$'. Null hypothesis is rejected in favour of hypothesis '$\text{left method} > \text{right method}$', if the corrected p-value is below a certain significance level. The method on the left side (of comparison) is placed on the x-axis, and the method on the right side is placed on the y-axis. Each cell represents the corrected p-value. Hence, every column represents the significance of the kappa score for a method when compared with other methods. Suppose the corrected p-value is less than a certain significance level in a cell. In that case, the null hypothesis is rejected, and the method on the x-axis will be assumed to have better performance than the corresponding method on the y-axis. The numerals in the x-axis represent the average Friedman ranking of the method.}
    \label{pavluechoenkappa}
\end{figure}

Finally, we analyze the statistical significance of the results. For this, we first subject the results to the Friedman ranking test. In the analysis, the average ranks of 3.07, 3.49, 3.66, 3.82, 3.94, 4.16, and 5.87 were obtained by GRAF, ET, OT, GB, XGB, RF, and ADA, respectively. With 114 datasets and 7 methods, the test statistic of the Friedman test was 117.6689. Assuming a significance level of $0.05$ with 6 degrees of freedom, the value of $\chi^{2}_{6}(0.05)=12.592$ is lesser than the test statistic. Hence, we reject the null hypothesis that all method's performances are similar. Now, we perform one-sided paired Wilcoxon signed-rank tests for every method to further demonstrate the statistical significance of the results. The six p-values for each method from the Wilcoxon test were corrected using the Bonferroni method~\cite{chen2017general}. Figure~\ref{pavluechoenkappa} shows that at a significance level of $0.05$, GRAF is significantly better than all other methods except for ET. Further, the methods have been arranged in increasing order of their Friedman ranking on the x-axis of Figure~\ref{pavluechoenkappa}.


\section{Sensitivity}\label{section::sensitivity}

We define the sensitivity of a region as the number of weights required to create it. It follows from the idea that regions with higher confusion will require more weights (hyperplanes) to purify them. We define a region with confusion as one in which samples of many different classes reside. We argue that points in these regions are crucial for approximating data, as these points have a major influence in defining the decision region.

We define the sensitivity of a point as a function of the number of weights required to put that sample into a pure region. To assign a sensitivity value to every point in a region, we first rank each point in the region arbitrarily and divide the sensitivity associated with a region by point's rank. Second, we normalize these values class-wise. If the region is big, ranked sensitivity prevents sensitivity scores from being overwhelmed with the points from a single region. On the other hand, class-wise normalization handles an imbalance in the data by assigning higher sensitivities to less populated classes. Formally, we represent the process as follows.

Let us assume, $v:\mathcal{F} \to \mathbb{N}$ maps each region to the number of weights required to pure it. Hence, the importance of each sample $x^{(i)}$ in the region $\Omega_p\in \mathcal{F}$ can be computed as

\begin{align}\label{eq::rankedSensitivity}
\theta_{p x^{(i)}} = \frac{v(\Omega_p)}{i}\:\:\forall i \in \{1,..,n_p\},\:\:1\leq p\leq P
\end{align}

Equation~\ref{eq::rankedSensitivity} assigns each sample in dataset an importance value, based on the size of region $\Omega_p$. Assume that the importance of a sample in dataset is given by $\theta_{x^{(i)}}$ $\forall i \in \{1,..,N\}$. Assuming that $X_{j}=\{x^{(k)}:y_{k}=j\:\:\forall k \in \{1,..,N\}\}$ $\forall j \in \{1,..,C\}$ represents a set of samples belonging to a class, the sensitivity of each sample can be computed as

\begin{align}
s_{i}=\ln\left(1+\frac{\theta_{x^{(i)}}}{\Theta_{y_{i}}}\right)\:\:\forall i \in \{1,..,N\},\text{ where } \Theta_{j} = \sum_{x^{(k)}\in X_{j}}\theta_{x^{(k)}},\:\:\forall j \in \{1,..,C\}
\end{align}

Assuming that each sample is assigned a sensitivity $s_{i}^{t}$ $\forall t \in \{1,..,T\}\land\forall i \in \{1,..,N\}$, the mean sensitivity of each sample can be defined as

\begin{align}
\hat{s}_{i} = \frac{1}{T}\sum_{t=1}^{T}s_{i}^{t}
\end{align}

Hence, the probability of each sample can be defined as

\begin{align}\label{eq::sensitivityProb}
p_{i}=\frac{\hat{s}_{i}}{\sum_{j=1}^{j=N}\hat{s}_{j}},\: \forall i \in \{1,..,N\}
\end{align}

The higher the probability or sensitivity of a sample, the more important it is.

\begin{figure}
    \centering
    \includegraphics[width=\linewidth,keepaspectratio]{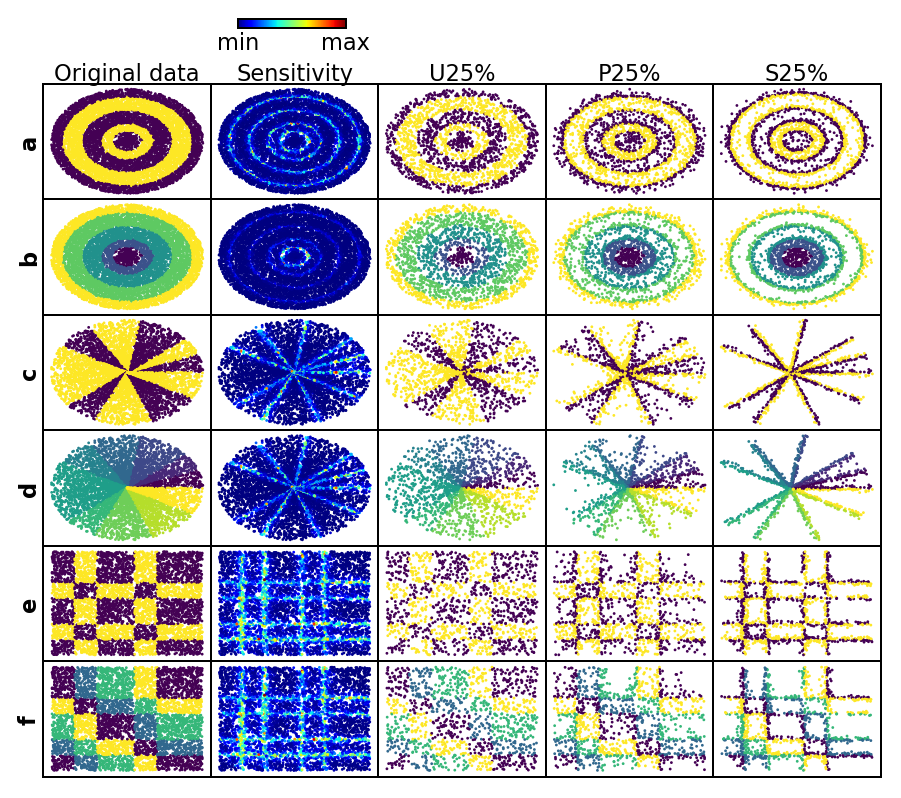}
    \caption{Assessment of performance of GRAF\'s sensitivity on simulated binary and multi-class datasets. (a, c, and e) represent simulated datasets with binary classes. (b, d, and f) represent simulated multi-class datasets. The classes are arranged in different patterns, concentric circles, pie-charts, and XOR representations, in a-b, c-d, and e-f, respectively. For each of these datasets, the distribution of sensitivities computed using GRAF has been shown in column \textit{Sensitivity}. A point with higher sensitivity indicates that it is more important for data approximation. The other columns U25\%, P25\%, and S25\%, compare the performances of data approximation using only 25\% of the total samples, sampled using a uniform distribution, distribution defined by GRAF\'s sensitivity, and the points with the highest values of sensitivities, respectively. The regions with the most confusion are best approximated using points with the highest sensitivities.   }
    \label{sensitivity}
\end{figure}

The sensitivities associated with the samples may be used to approximate the complete dataset, for further downstream analyses with high sensitivity points only. A study was designed to assess how well the sensitivity computed using GRAF approximates different datasets. To perform this analysis, 6 different datasets were created. Every dataset consists of samples distributed in different patterns (concentric circles, pie-charts, and XOR representations). For every pattern, both binary and multi-class versions were generated, as illustrated in Figure~\ref{sensitivity}. To generate sensitivity scores on each dataset, 200 trees ($L=200$) with complete features space ($M=2$) were generated and sensitivity score ($\hat{s}_{i}$) was computed. The performance of GRAF\'s sensitivity has been compared with a uniform distribution for samples. Figure~\ref{sensitivity} illustrates that when only 25\% of the total points are sampled, samples with the highest sensitivities adequately approximate the regions with the highest confusion. 

\begin{figure}
    \centering
    \includegraphics[width=\linewidth,keepaspectratio]{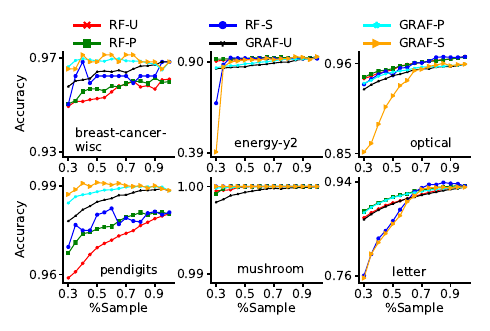}
    \caption{Performance evaluation of Random forest~\cite{breiman2001random} and GRAF, with increasing fraction of samples used for training, sampled according to uniform distribution (U), their sensitivities (P), and their decreasing order of sensitivities (S). The points sampled using distribution defined by their sensitivities perform comparable or better when compared with points sampled using uniform distribution. Also, as points are added in the decreasing order of their sensitivities, the accuracy on test set converges and reaches its maximum with only a fraction of points with high sensitivities. The trends in results are similar, irrespective of the method used for classification. }
    \label{combined_acc}
\end{figure}

If points are sampled from two different distributions- 1. uniform, 2. distribution defined by sensitivities associated with points, then the performance of the latter is better than former (Figure~\ref{combined_acc}). Further, the maximal accuracy on a test set can be achieved by using only a fraction of its samples with the highest sensitivities (Figure~\ref{combined_acc}). Similar trends in results are observed, irrespective of the method (Random forest~\cite{breiman2001random} or GRAF) used for learning the model. This study also enforces the idea that high sensitivity points approximate the decision boundary reasonably well. To perform this experiment, 200 trees ($L=200$) were generated, and the number of features ($M$) was chosen as per the tuned model, and sensitivity scores were computed on the resulting trees.

\begin{figure}
    \centering
    \includegraphics[width=\linewidth,keepaspectratio]{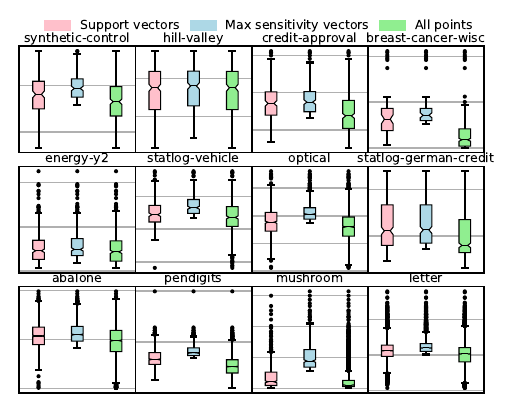}
    \caption{An analogy between support vectors and points with high sensitivities. The distribution of probabilities (\ref{eq::sensitivityProb}) associated with support vectors has been compared with that of a fraction of points with high sensitivities, and the distribution of probabilities associated with all points. It can be concluded that points with higher sensitivities coincide with the support vectors with higher values of weights.}
    \label{combined_box}
\end{figure}

The extension of the previous study has been done to show that high sensitivity points found by GRAF are analogous to support vectors. The performance of GRAF is compared with two well-known methods used for reducing the samples in training set for SVM~\cite{nalepa2019selecting}. Neighborhood Property-Based Pattern Selection (NPPS)~\cite{shin2007neighborhood} selects points near the decision boundary by utilizing the property that \textit{"a pattern located near the decision boundary tends to have more heterogeneous neighbors in its class membership"}. A sample has a heterogeneous neighborhood when a few of its immediate neighbors belong to different classes. The measure for the heterogeneity in the neighborhood of a point is given by (negative) entropy. For points with high heterogeneity (high entropy) in their neighborhood, they are selected from the training set. The performance of NPPS algorithm heavily depends on the initial value of the number of clusters $k$. Thus, in the experiments, the value of $k$ was tuned from $2$ to $50$. The reduced set corresponding to that $k$ for which the SVM model had the highest performance on the test data was selected for comparison. The second method for comparison is an ensemble method called Small Votes Instance Selection (SVIS)~\cite{guo2015fast}. SVIS selects points with small values of ensemble margin (\ref{eq::treeMargin}). A sample with a small margin tends to lie near the decision boundary, and hence, is more informative to build a classifier. In the experiments, an ensemble of 100 decision trees with bagging was created. As suggested by authors~\cite{guo2015fast}, the different bags of datasets were generated by sampling (with replacement) $63.2\%$ of samples from the training set.  

Table~\ref{tab:supportvector} records the accuracy on a given test set when an SVM model was trained using all the samples in the training set. These results were compared with an SVM model that is trained using only the high sensitivity points of GRAF, the points with a low margin in SVIS, and the reduced training set of NPPS. The size of the reduced training set for GRAF and SVIS was chosen such that it constituted the same fraction as that of support vectors (SVs). An analogy between support vectors and the fraction of points with high sensitivity points from GRAF has also been illustrated in Figure~\ref{combined_box}. The SVM's performance on the reduced training set selected by all three methods is almost similar and is in very close proximity to SVM's performance when trained on the complete training set.

\begin{table*}[!ht]
\makebox[1 \linewidth][c]{
\resizebox{1.1 \linewidth}{!}{ %
\begin{tabular}{||lrrr|rr|rr|rrrr||}
\toprule
                        &           &       &           & GRAF      &               & SVIS      &               &       &               &NPPS           &               \\
\cmidrule{5-12}
                        &           &       &           &           &\%SVM          &           &\%SVM          &       &               &           &\%SVM          \\
                        &           &       &           &           &accuracy on    &           &accuracy on    &       &\%size of      &           &accuracy on    \\
                        &\#Train    &       &\%SVM      &\%Overlap  &reduced        &\%Overlap  &reduced        &       &reduced        &\%Overlap  &reduced        \\
Dataset                 &Samples    &\%SVs  &Accuracy   &with SVs   &training set   &with SVs   &training set   &k      &training set   &with SVs   &training set   \\
\midrule
synthetic-control       &300        &55.67  &99.00      &67.67      &98.00          &61.08      &94.00          &21     &59.00          &70.06      &99.00\\

hill-valley             &303        &95.71  &49.83      &95.52      &50.17          &95.86      &50.83          &49     &59.41          &57.93      &52.48\\

credit-approval         &345        &53.62  &87.54      &74.59      &87.54          &78.92      &87.54          &26     &79.13          &64.86      &88.12\\

breast-cancer-wisc      &350        &17.43  &96.85      &55.74      &96.56          &65.57      &96.28          &30     &20.00          &37.70      &96.56\\

energy-y2               &384        &80.73  &90.89      &84.52      &90.89          &83.87      &90.10          &50     &61.98          &65.81      &71.35\\

statlog-vehicle         &423        &52.72  &79.91      &58.30      &79.91          &79.37      &68.56          &28     &85.34          &74.89      &79.67\\

statlog-german-credit   &500        &60.80  &74.00      &87.83      &75.80          &84.87      &74.80          &7      &51             &53.62      &73.40\\

titanic                 &1101       &43.32  &78.64      &39.83      &78.64          &50.73      &65.09          &46     &24.34          &16.14      &78.64\\

optical                 &1912       &39.33  &98.33      &60.51      &97.75          &66.09      &96.81          &49     &69.61          &90.82      &98.38\\

abalone                 &2089       &68.12  &66.14      &83.91      &64.85          &86.16      &48.75          &7      &59.65          &65.14      &66.04\\

pendigits               &3747       &19.51  &99.52      &51.30      &99.20          &51.85      &95.92          &45     &32.99          &70.59      &97.57\\

mushroom                &4062       &11.18  &100.00     &27.75      &100.00         &21.37      &50.76          &45     &5.15           &15.86      &78.75\\

letter                  &10000      &52.19  &96.53      &67.89      &92.34          &71.60      &92.49          &50     &85.64          &95.65      &96.41\\

\bottomrule
\end{tabular}
}}
\caption{Equivalence between the reduced training set and support vectors. For a given test set, the SVM model is learned using two different sets. First, an SVM model is trained using all the samples in the training set. Its accuracy on the test set is then evaluated (column \textit{\% SVM Accuracy}), and  information about the support vectors is recorded (column \textit{\% SVs}). Separately, an SVM model is trained using points from the reduced training set (column \textit{\% SVM accuracy on reduced training set}). For GRAF and SVIS, the size of the reduced training set is the same as that of support vectors. For NPPS, the reduced training set consists of samples with high heterogeneity values in their neighborhood (column \textit{\%size of reduced training set}). The size of the neighborhood in NPPS is determined by $k$. An analogy between the reduced training set and support vectors is recorded in column \textit{\% Overlap with SVs}, for all three methods. Note that the hyper-parameters for the SVM model in the reduced training set were kept the same as that of the full training set.}
\label{tab:supportvector}
\end{table*}

\section{Conclusion}\label{section::conclusion}
In this paper, we propose a supervised approach to constructing random forests, termed as Guided Random Forest (GRAF). GRAF repeatedly draws random hyperplanes to partition the data. It uses successive hyperplanes to correct impure partitions to the extent feasible, so that the overall purity of resultant partitions increases. The resultant partitions (or leaf nodes) are represented with variable length codes. This guided tree construction bridges the gap between boosting and decision trees, where every tree represents a high variance instance. Results on 115 benchmark datasets show that GRAF outperforms state of the art bagging and boosting based algorithms like Random Forest~\cite{breiman2001random} and Gradient Boosting~\cite{friedman2001greedy}. The results show that GRAF is effective on both binary and multi-class datasets. GRAF exhibits both low bias and low variance with increasing size of the training dataset. We introduce the notion of sensitivity, a metric that indicates the importance of a sample. We show that GRAF can be used to approximate a given dataset by using only a few high sensitivity points. The proposed sensitivity concept does not dwell into the selection criteria for a subset of points. However, it differentiates between points on the basis of their proximity to confusion regions, akin to support vectors in kernel schemes.

\appendix
\section*{Appendices}

\renewcommand{\thefigure}{A.\arabic{figure}}
\renewcommand{\thetable}{A.\arabic{table}}
\renewcommand{\thealgorithm}{A.\arabic{algorithm}}
\setcounter{figure}{0} 
\setcounter{table}{0}
\setcounter{algorithm}{0}
\section{Data generation with Weka for Bias-variance tradeoff}\label{sec:dataGeneration}
In order to examine bias-variance tradeoff, 6 different binary and multi-class datasets with different number of centroids were generated using Weka~\cite{witten1999weka}. The RandomRBF data generator was selected to simulate the data. A detailed description of this class is available at \url{http://weka.sourceforge.net/doc.dev/weka/datagenerators/classifiers/classification/RandomRBF.html}. In order to generate the data set, the number of features '-a' was set to 10, the number of centroids '-C' was selected from \{10, 20, 50\}, and the number of classes '-c' was selected from \{2, 5\}. For each dataset, a total of 10000 samples '-n' were generated. The commands to generate the data from weka with seed '-S' 1 are given below:

\begin{verbatim}
java -Xmx128m -classpath $PWD:weka.jar weka.datagenerators.classifiers.
classification.RandomRBF -r weka.datagenerators.classifiers.classification.
RandomRBF-datafile -S 1 -n 10000 -a 10 -c 2 -C 10
\end{verbatim}

\begin{verbatim}
java -Xmx128m -classpath $PWD:weka.jar weka.datagenerators.classifiers.
classification.RandomRBF -r weka.datagenerators.classifiers.classification.
RandomRBF-datafile -S 1 -n 10000 -a 10 -c 5 -C 10
\end{verbatim}

\begin{verbatim}
java -Xmx128m -classpath $PWD:weka.jar weka.datagenerators.classifiers.
classification.RandomRBF -r weka.datagenerators.classifiers.classification.
RandomRBF-datafile -S 1 -n 10000 -a 10 -c 2 -C 20
\end{verbatim}

\begin{verbatim}
java -Xmx128m -classpath $PWD:weka.jar weka.datagenerators.classifiers.
classification.RandomRBF -r weka.datagenerators.classifiers.classification.
RandomRBF-datafile -S 1 -n 10000 -a 10 -c 5 -C 20
\end{verbatim}

\begin{verbatim}
java -Xmx128m -classpath $PWD:weka.jar weka.datagenerators.classifiers.
classification.RandomRBF -r weka.datagenerators.classifiers.classification.
RandomRBF-datafile -S 1 -n 10000 -a 10 -c 2 -C 50
\end{verbatim}

\begin{verbatim}
java -Xmx128m -classpath $PWD:weka.jar weka.datagenerators.classifiers.
classification.RandomRBF -r weka.datagenerators.classifiers.classification.
RandomRBF-datafile -S 1 -n 10000 -a 10 -c 5 -C 50
\end{verbatim}

\renewcommand{\thefigure}{B.\arabic{figure}}
\renewcommand{\thetable}{B.\arabic{table}}
\renewcommand{\thealgorithm}{B.\arabic{algorithm}}
\setcounter{figure}{0} 
\setcounter{table}{0}
\setcounter{algorithm}{0}
\section{More results on Bias-variance tradeoff}
\begin{figure}[!ht]
	\centering
	\includegraphics[width=\linewidth,keepaspectratio]{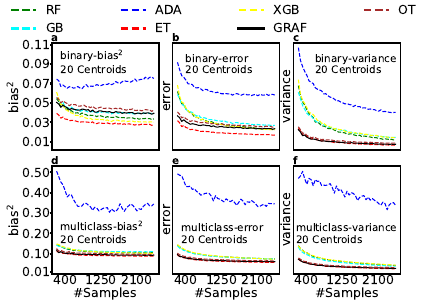}
	\caption{Bias-variance analysis with increasing \#samples in a training set. For both (a-c) binary and (d-f) multi-class datasets with 20 centroids, the number of training samples is increased from $200$ to $2500$, while keeping number of features to be sampled fixed at $(M = n/2)$, and the number of estimators kept at ($L=100$). As the cardinality of the training set is increased, bias and variance continue to decrease, and misclassification error continues to decrease and may asymptotically reach its minimum.}
	\label{comparison20}
\end{figure}

\begin{figure}[!ht]
	\centering
	\includegraphics[width=\linewidth,keepaspectratio]{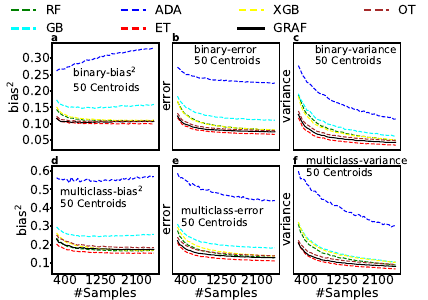}
	\caption{Bias-variance analysis with increasing \#samples in a training set. For both (a-c) binary and (d-f) multi-class datasets with 50 centroids, the number of training samples is increased from $200$ to $2500$, while fixing the number of features to be sampled at $(M = n/2)$, and the number of estimators at ($L=100$). As the cardinality of the training set is increased, bias and variance continues to increase, and the misclassification error continues to decrease and may saturate to its minimum.}
	\label{comparison50}
\end{figure}

\begin{figure}[!ht]
    \centering
    \includegraphics[width=\linewidth,keepaspectratio]{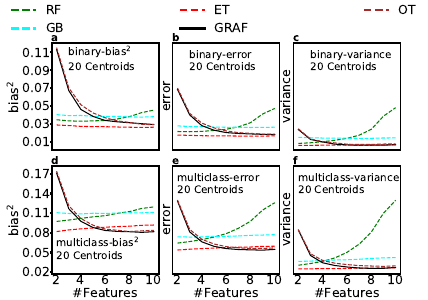}
    \caption{Bias-variance analysis with an increasing number of dimensions (features) selected from a given feature space in a classifier. For both binary (a-c) and multi-class (d-f) datasets with 20 centroids, $M$  is increased from $2$ to $10$, while fixing the number of estimators to be ensembled $(L = 100)$. For GRAF, when the dimension of the sub-space is large enough to distinguish samples of different classes, bias and variance saturate and converge to their minimum. With increasing dimensionality of the sub-space, misclassification error continues to decrease and rapidly saturates to its minimum.}
    \label{increasingM20}
\end{figure}

\begin{figure}[!ht]
    \centering
    \includegraphics[width=\linewidth,keepaspectratio]{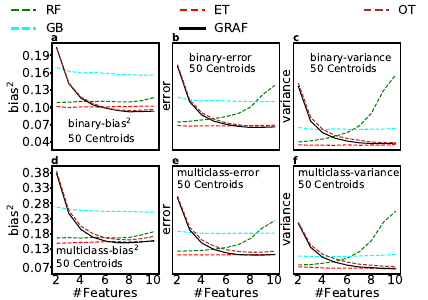}
    \caption{Bias-variance analysis with an increasing number of dimensions (features) selected from a given feature space in a classifier. For both binary (a-c) and multi-class (d-f) datasets with 50 centroids, $M$  is increased from $2$ to $10$, while fixing the number of estimators to be ensembled $(L = 100)$. For GRAF, when the dimension of the sub-space is large enough to distinguish samples of different classes, bias and variance saturate and converge to their minimum. With increasing dimensionality of the sub-space, misclassification error continues to decrease and rapidly saturates to its minimum.}
    \label{increasingM50}
\end{figure}

\begin{figure}[!ht]
    \centering
    \includegraphics[width=\linewidth,keepaspectratio]{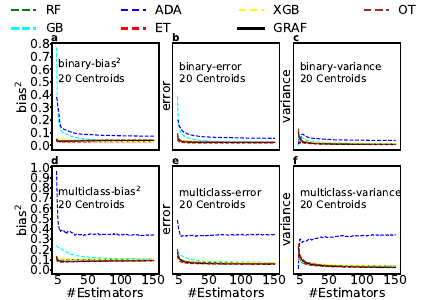}
    \caption{Bias-variance analysis with an increasing number of estimators (trees) in a classifier. For both binary (a-c) and multi-class (d-f) datasets with 20 centroids, the number of estimators is increased from $2$ to $150$, while fixing the number of dimensions to be sampled $(M = n/2)$. As the number of estimators is increased, bias, error, and variance rapidly saturate.}
    \label{increasingL20}
\end{figure}

\begin{figure}[!ht]
    \centering
    \includegraphics[width=\linewidth,keepaspectratio]{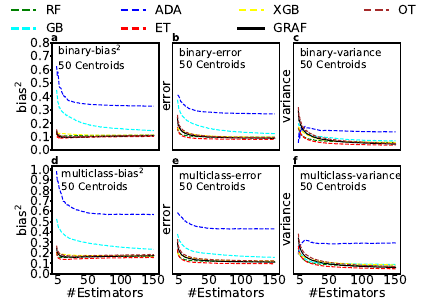}
    \caption{Bias-variance analysis with an increasing number of estimators (trees) in a classifier. For both binary (a-c) and multi-class (d-f) datasets with 50 centroids, the number of estimators is increased from $2$ to $150$, while fixing the number of dimensions to be sampled $(M = n/2)$. As the number of estimators is increased, bias, error, and variance rapidly saturate.}
    \label{increasingL50}
\end{figure}

\renewcommand{\thefigure}{C.\arabic{figure}}
\renewcommand{\thetable}{C.\arabic{table}}
\renewcommand{\thealgorithm}{C.\arabic{algorithm}}
\setcounter{figure}{0} 
\setcounter{table}{0}
\setcounter{algorithm}{0}
\begin{landscape}
\section{Data statistics of UCI datasets}
\begin{center}
\singlespacing
\begin{longtable}{||c|c|c|c|c|c|c|c||}
\hline
Dataset & nFeatures & nClasses & nSamples & imbalance & PC(v=0.5) & PC(v=0.7) & PC(v=0.9) \\ [0.5ex] 
\hline\hline
abalone&8&3&4177&N&1.0&1.0&2.0\\
\hline
acute-inflammation&6&2&120&N&2.0&3.0&4.0\\
\hline
acute-nephritis&6&2&120&Y&2.0&3.0&4.0\\
\hline
adult&14&2&32561&Y&6.0&9.0&12.0\\
\hline
arrhythmia&262&13&452&Y&12.0&25.0&55.0\\
\hline
audiology-std&59&18&171&Y&10.0&16.0&26.0\\
\hline
balance-scale&4&3&625&Y&2.0&3.0&4.0\\
\hline
bank&16&2&4521&Y&6.0&9.0&13.0\\
\hline
blood&4&2&748&Y&1.0&2.0&2.0\\
\hline
breast-cancer&9&2&286&Y&3.0&5.0&7.0\\
\hline
breast-cancer-wisc&9&2&699&Y&1.0&2.0&6.0\\
\hline
breast-cancer-wisc-diag&30&2&569&Y&2.0&3.0&7.0\\
\hline
breast-cancer-wisc-prog&33&2&198&Y&2.0&4.0&9.0\\
\hline
breast-tissue&9&6&106&Y&1.0&2.0&3.0\\
\hline
car&6&4&1728&Y&3.0&5.0&6.0\\
\hline
cardiotocography-10clases&21&10&2126&Y&3.0&6.0&11.0\\
\hline
cardiotocography-3clases&21&3&2126&Y&3.0&6.0&11.0\\
\hline
chess-krvk&6&18&28056&Y&3.0&4.0&5.0\\
\hline
chess-krvkp&36&2&3196&N&9.0&16.0&26.0\\
\hline
congressional-voting&16&2&435&Y&4.0&7.0&11.0\\
\hline
conn-bench-sonar-mines-rocks&60&2&208&N&4.0&8.0&20.0\\
\hline
conn-bench-vowel-deterding&11&11&528&N&3.0&4.0&7.0\\
\hline
connect-4&42&2&67557&Y&9.0&17.0&31.0\\
\hline
contrac&9&3&1473&Y&3.0&5.0&7.0\\
\hline
credit-approval&15&2&690&Y&4.0&7.0&11.0\\
\hline
cylinder-bands&35&2&512&Y&6.0&12.0&21.0\\
\hline
dermatology&34&6&366&Y&3.0&8.0&16.0\\
\hline
echocardiogram&10&2&131&Y&3.0&5.0&7.0\\
\hline
energy-y1&8&3&768&Y&2.0&3.0&5.0\\
\hline
energy-y2&8&3&768&Y&2.0&3.0&5.0\\
\hline
fertility&9&2&100&Y&3.0&5.0&7.0\\
\hline
glass&9&6&214&Y&2.0&3.0&5.0\\
\hline
haberman-survival&3&2&306&Y&2.0&2.0&3.0\\
\hline
hayes-roth&3&3&132&Y&2.0&2.0&3.0\\
\hline
heart-cleveland&13&5&303&Y&4.0&6.0&10.0\\
\hline
heart-hungarian&12&2&294&Y&3.0&6.0&9.0\\
\hline
heart-switzerland&12&5&123&Y&4.0&6.0&9.0\\
\hline
heart-va&12&5&200&Y&3.0&5.0&8.0\\
\hline
hepatitis&19&2&155&Y&4.0&7.0&13.0\\
\hline
hill-valley&100&2&606&N&1.0&1.0&1.0\\
\hline
horse-colic&25&2&300&Y&5.0&10.0&18.0\\
\hline
ilpd-indian-liver&9&2&583&Y&2.0&3.0&5.0\\
\hline
image-segmentation&18&7&210&N&1.0&3.0&6.0\\
\hline
ionosphere&33&2&351&Y&4.0&8.0&16.0\\
\hline
iris&4&3&150&N&1.0&1.0&2.0\\
\hline
led-display&7&10&1000&Y&3.0&4.0&6.0\\
\hline
lenses&4&3&24&Y&2.0&3.0&4.0\\
\hline
letter&16&26&20000&N&3.0&6.0&10.0\\
\hline
libras&90&15&360&N&3.0&4.0&7.0\\
\hline
low-res-spect&100&9&531&Y&1.0&2.0&4.0\\
\hline
lung-cancer&56&3&32&Y&4.0&7.0&11.0\\
\hline
lymphography&18&4&148&Y&4.0&7.0&12.0\\
\hline
magic&10&2&19020&Y&2.0&4.0&6.0\\
\hline
mammographic&5&2&961&Y&2.0&3.0&4.0\\
\hline
miniboone&50&2&130064&Y&1.0&1.0&3.0\\
\hline
molec-biol-promoter&57&2&106&N&10.0&16.0&27.0\\
\hline
molec-biol-splice&60&3&3190&Y&24.0&37.0&51.0\\
\hline
monks-1&6&2&124&N&3.0&4.0&6.0\\
\hline
monks-2&6&2&169&Y&3.0&4.0&6.0\\
\hline
monks-3&6&2&122&N&3.0&4.0&5.0\\
\hline
mushroom&21&2&8124&N&4.0&7.0&13.0\\
\hline
musk-1&166&2&476&Y&3.0&7.0&23.0\\
\hline
musk-2&166&2&6598&Y&3.0&9.0&26.0\\
\hline
nursery&8&5&12960&Y&4.0&6.0&8.0\\
\hline
oocytes\_merluccius\_nucleus\_4d&41&2&1022&Y&1.0&1.0&3.0\\
\hline
oocytes\_merluccius\_states\_2f&25&3&1022&Y&2.0&3.0&5.0\\
\hline
oocytes\_trisopterus\_nucleus\_2f&25&2&912&Y&2.0&3.0&5.0\\
\hline
oocytes\_trisopterus\_states\_5b&32&3&912&Y&1.0&2.0&5.0\\
\hline
optical&62&10&3823&N&8.0&15.0&30.0\\
\hline
ozone&72&2&2536&Y&2.0&4.0&12.0\\
\hline
page-blocks&10&5&5473&Y&2.0&3.0&5.0\\
\hline
parkinsons&22&2&195&Y&1.0&2.0&6.0\\
\hline
pendigits&16&10&7494&N&3.0&4.0&8.0\\
\hline
pima&8&2&768&Y&3.0&4.0&6.0\\
\hline
pittsburg-bridges-MATERIAL&7&3&106&Y&3.0&4.0&6.0\\
\hline
pittsburg-bridges-REL-L&7&3&103&Y&2.0&4.0&6.0\\
\hline
pittsburg-bridges-SPAN&7&3&92&Y&2.0&4.0&6.0\\
\hline
pittsburg-bridges-T-OR-D&7&2&102&Y&3.0&4.0&6.0\\
\hline
pittsburg-bridges-TYPE&7&6&105&Y&2.0&4.0&6.0\\
\hline
planning&12&2&182&Y&3.0&4.0&5.0\\
\hline
plant-margin&64&100&1600&N&4.0&8.0&25.0\\
\hline
plant-shape&64&100&1600&N&1.0&1.0&2.0\\
\hline
plant-texture&64&100&1599&N&6.0&13.0&30.0\\
\hline
post-operative&8&3&90&Y&3.0&4.0&6.0\\
\hline
ringnorm&20&2&7400&N&10.0&14.0&18.0\\
\hline
seeds&7&3&210&N&1.0&1.0&3.0\\
\hline
semeion&256&10&1593&N&16.0&36.0&103.0\\
\hline
soybean&35&18&307&Y&5.0&10.0&19.0\\
\hline
spambase&57&2&4601&Y&15.0&26.0&41.0\\
\hline
spect&22&2&79&Y&3.0&6.0&11.0\\
\hline
spectf&44&2&80&N&2.0&3.0&10.0\\
\hline
statlog-australian-credit&14&2&690&Y&4.0&7.0&10.0\\
\hline
statlog-german-credit&24&2&1000&Y&7.0&11.0&18.0\\
\hline
statlog-heart&13&2&270&Y&4.0&6.0&10.0\\
\hline
statlog-image&18&7&2310&N&2.0&4.0&8.0\\
\hline
statlog-landsat&36&6&4435&Y&2.0&2.0&4.0\\
\hline
statlog-shuttle&9&7&43500&Y&3.0&4.0&6.0\\
\hline
statlog-vehicle&18&4&846&N&1.0&2.0&5.0\\
\hline
steel-plates&27&7&1941&Y&3.0&5.0&10.0\\
\hline
synthetic-control&60&6&600&N&1.0&4.0&18.0\\
\hline
teaching&5&3&151&N&2.0&3.0&5.0\\
\hline
thyroid&21&3&3772&Y&7.0&11.0&16.0\\
\hline
tic-tac-toe&9&2&958&Y&4.0&5.0&7.0\\
\hline
titanic&3&2&2201&Y&2.0&2.0&3.0\\
\hline
twonorm&20&2&7400&N&8.0&13.0&18.0\\
\hline
vertebral-column-2clases&6&2&310&Y&1.0&2.0&4.0\\
\hline
vertebral-column-3clases&6&3&310&Y&1.0&2.0&3.0\\
\hline
wall-following&24&4&5456&Y&5.0&10.0&18.0\\
\hline
waveform&21&3&5000&N&2.0&6.0&15.0\\
\hline
waveform-noise&40&3&5000&N&10.0&19.0&29.0\\
\hline
wine&13&3&178&Y&2.0&4.0&7.0\\
\hline
wine-quality-red&11&6&1599&Y&3.0&4.0&7.0\\
\hline
wine-quality-white&11&7&4898&Y&3.0&5.0&8.0\\
\hline
yeast&8&10&1484&Y&3.0&5.0&7.0\\
\hline
zoo&16&7&101&Y&2.0&4.0&8.0\\
\hline
\caption{Data statistics of 115 UCI datasets. The total number of samples across all datasets varies from 24 to $\sim$130k. The count of features across all datasets varies from 3 to 262.}
\label{dataStat}
\end{longtable}
\end{center}

\renewcommand{\thefigure}{D.\arabic{figure}}
\renewcommand{\thetable}{D.\arabic{table}}
\renewcommand{\thealgorithm}{D.\arabic{algorithm}}
\setcounter{figure}{0} 
\setcounter{table}{0}
\setcounter{algorithm}{0}
\section{Results on UCI datasets}
\subsection*{Cohen's Kappa Coefficient}
\begin{center}
\singlespacing
\begin{longtable}{||c|c|c|c|c|c|c|c||}
\hline
Dataset & GRAF & OT & ET & GB & ADA & RF & XGB \\
\hline\hline
abalone&0.488$\pm$0.005&\textbf{0.492$\pm$0.016}&0.484$\pm$0.009&0.469$\pm$0.007&0.458$\pm$0.008&0.483$\pm$0.016&0.466$\pm$0.013\\
\hline
acute-inflammation&\textbf{1.000$\pm$0.000}&\textbf{1.000$\pm$0.000}&\textbf{1.000$\pm$0.000}&\textbf{1.000$\pm$0.000}&\textbf{1.000$\pm$0.000}&\textbf{1.000$\pm$0.000}&\textbf{1.000$\pm$0.000}\\
\hline
acute-nephritis&\textbf{1.000$\pm$0.000}&\textbf{1.000$\pm$0.000}&\textbf{1.000$\pm$0.000}&\textbf{1.000$\pm$0.000}&\textbf{1.000$\pm$0.000}&\textbf{1.000$\pm$0.000}&\textbf{1.000$\pm$0.000}\\
\hline
adult&0.602$\pm$0.005&0.600$\pm$0.004&0.571$\pm$0.005&\textbf{0.631$\pm$0.003}&0.625$\pm$0.004&0.594$\pm$0.002&0.630$\pm$0.004\\
\hline
arrhythmia&0.403$\pm$0.058&0.342$\pm$0.026&\textbf{0.628$\pm$0.046}&0.567$\pm$0.023&0.258$\pm$0.049&0.614$\pm$0.032&0.582$\pm$0.020\\
\hline
audiology-std&\textbf{0.843$\pm$0.039}&0.741$\pm$0.061&0.785$\pm$0.075&0.800$\pm$0.064&0.605$\pm$0.071&0.785$\pm$0.083&0.792$\pm$0.071\\
\hline
balance-scale&0.842$\pm$0.027&0.850$\pm$0.029&0.755$\pm$0.021&0.860$\pm$0.044&\textbf{0.887$\pm$0.028}&0.763$\pm$0.028&0.807$\pm$0.022\\
\hline
bank&\textbf{0.465$\pm$0.052}&0.382$\pm$0.040&0.317$\pm$0.048&0.391$\pm$0.018&0.338$\pm$0.034&0.391$\pm$0.057&0.405$\pm$0.023\\
\hline
blood&0.265$\pm$0.029&\textbf{0.291$\pm$0.044}&0.233$\pm$0.085&0.238$\pm$0.078&0.089$\pm$0.008&0.233$\pm$0.089&0.212$\pm$0.068\\
\hline
breast-cancer&\textbf{0.438$\pm$0.037}&0.433$\pm$0.043&0.353$\pm$0.059&0.278$\pm$0.134&0.283$\pm$0.101&0.336$\pm$0.079&0.319$\pm$0.135\\
\hline
breast-cancer-wisc&0.953$\pm$0.016&\textbf{0.956$\pm$0.018}&0.950$\pm$0.015&0.931$\pm$0.020&0.944$\pm$0.018&0.947$\pm$0.013&0.934$\pm$0.028\\
\hline
breast-cancer-wisc-diag&\textbf{0.947$\pm$0.023}&0.935$\pm$0.027&0.928$\pm$0.027&0.928$\pm$0.023&0.920$\pm$0.020&0.909$\pm$0.018&0.936$\pm$0.023\\
\hline
breast-cancer-wisc-prog&\textbf{0.431$\pm$0.042}&0.418$\pm$0.048&0.325$\pm$0.071&0.322$\pm$0.121&0.212$\pm$0.211&0.263$\pm$0.153&0.241$\pm$0.172\\
\hline
breast-tissue&0.706$\pm$0.085&\textbf{0.718$\pm$0.067}&0.647$\pm$0.097&0.590$\pm$0.134&0.451$\pm$0.061&0.684$\pm$0.111&0.613$\pm$0.134\\
\hline
car&0.941$\pm$0.020&0.922$\pm$0.012&0.968$\pm$0.009&0.986$\pm$0.013&0.719$\pm$0.008&0.975$\pm$0.006&\textbf{0.987$\pm$0.009}\\
\hline
cardiotocography-10clases&0.800$\pm$0.015&0.800$\pm$0.011&0.841$\pm$0.017&0.868$\pm$0.011&0.637$\pm$0.038&0.841$\pm$0.012&\textbf{0.874$\pm$0.011}\\
\hline
cardiotocography-3clases&0.791$\pm$0.031&0.773$\pm$0.024&0.867$\pm$0.018&\textbf{0.885$\pm$0.023}&0.709$\pm$0.014&0.848$\pm$0.022&0.878$\pm$0.025\\
\hline
chess-krvk&0.694$\pm$0.002&0.626$\pm$0.003&0.858$\pm$0.004&\textbf{0.911$\pm$0.001}&0.120$\pm$0.006&0.855$\pm$0.003&0.907$\pm$0.002\\
\hline
chess-krvkp&0.955$\pm$0.017&0.953$\pm$0.016&\textbf{0.994$\pm$0.004}&0.993$\pm$0.003&0.940$\pm$0.011&0.990$\pm$0.007&0.991$\pm$0.004\\
\hline
congressional-voting&0.212$\pm$0.032&\textbf{0.215$\pm$0.036}&0.003$\pm$0.033&0.048$\pm$0.064&0.031$\pm$0.039&0.030$\pm$0.042&0.035$\pm$0.046\\
\hline
conn-bench-sonar-mines-rocks&0.697$\pm$0.070&0.667$\pm$0.020&\textbf{0.765$\pm$0.049}&0.605$\pm$0.049&0.582$\pm$0.055&0.608$\pm$0.088&0.747$\pm$0.088\\
\hline
conn-bench-vowel-deterding&0.975$\pm$0.013&0.975$\pm$0.012&\textbf{0.979$\pm$0.018}&0.944$\pm$0.012&0.562$\pm$0.019&0.958$\pm$0.018&0.871$\pm$0.022\\
\hline
connect-4&0.677$\pm$0.003&0.678$\pm$0.004&0.663$\pm$0.002&0.750$\pm$0.003&0.499$\pm$0.003&0.620$\pm$0.002&\textbf{0.763$\pm$0.005}\\
\hline
contrac&0.275$\pm$0.013&0.273$\pm$0.022&0.242$\pm$0.035&0.294$\pm$0.047&0.281$\pm$0.026&0.261$\pm$0.040&\textbf{0.302$\pm$0.050}\\
\hline
credit-approval&\textbf{0.784$\pm$0.048}&0.769$\pm$0.039&0.720$\pm$0.071&0.768$\pm$0.052&0.717$\pm$0.048&0.754$\pm$0.017&0.738$\pm$0.042\\
\hline
cylinder-bands&0.560$\pm$0.022&0.557$\pm$0.034&0.594$\pm$0.026&0.601$\pm$0.062&0.489$\pm$0.067&0.583$\pm$0.051&\textbf{0.642$\pm$0.046}\\
\hline
dermatology&\textbf{0.976$\pm$0.011}&\textbf{0.976$\pm$0.011}&\textbf{0.976$\pm$0.006}&0.969$\pm$0.006&0.917$\pm$0.022&0.972$\pm$0.010&0.962$\pm$0.006\\
\hline
echocardiogram&\textbf{0.630$\pm$0.110}&0.603$\pm$0.048&0.579$\pm$0.023&0.524$\pm$0.098&0.606$\pm$0.041&0.578$\pm$0.038&0.518$\pm$0.154\\
\hline
energy-y1&0.912$\pm$0.015&0.908$\pm$0.019&0.945$\pm$0.030&0.935$\pm$0.011&0.682$\pm$0.010&\textbf{0.950$\pm$0.006}&0.945$\pm$0.004\\
\hline
energy-y2&0.848$\pm$0.015&0.852$\pm$0.013&0.835$\pm$0.031&\textbf{0.871$\pm$0.021}&0.785$\pm$0.007&0.842$\pm$0.020&0.846$\pm$0.018\\
\hline
fertility&0.335$\pm$0.230&\textbf{0.351$\pm$0.203}&0.218$\pm$0.251&0.285$\pm$0.358&0.000$\pm$0.000&0.201$\pm$0.206&0.229$\pm$0.138\\
\hline
glass&\textbf{0.737$\pm$0.113}&0.718$\pm$0.109&0.679$\pm$0.049&0.667$\pm$0.054&0.462$\pm$0.074&0.699$\pm$0.040&0.689$\pm$0.073\\
\hline
haberman-survival&\textbf{0.249$\pm$0.106}&0.235$\pm$0.101&0.091$\pm$0.024&0.098$\pm$0.054&0.172$\pm$0.107&0.049$\pm$0.060&0.241$\pm$0.054\\
\hline
hayes-roth&0.754$\pm$0.050&0.754$\pm$0.050&0.742$\pm$0.069&0.741$\pm$0.072&\textbf{0.778$\pm$0.062}&0.754$\pm$0.052&0.765$\pm$0.090\\
\hline
heart-cleveland&\textbf{0.323$\pm$0.051}&0.295$\pm$0.050&0.304$\pm$0.027&0.279$\pm$0.089&0.245$\pm$0.060&0.285$\pm$0.067&0.256$\pm$0.065\\
\hline
heart-hungarian&\textbf{0.696$\pm$0.066}&0.686$\pm$0.068&0.653$\pm$0.063&0.613$\pm$0.074&0.624$\pm$0.037&0.653$\pm$0.070&0.597$\pm$0.042\\
\hline
heart-switzerland&0.250$\pm$0.061&\textbf{0.253$\pm$0.107}&0.121$\pm$0.033&0.088$\pm$0.054&0.134$\pm$0.105&0.132$\pm$0.094&0.096$\pm$0.048\\
\hline
heart-va&\textbf{0.175$\pm$0.040}&0.148$\pm$0.061&0.073$\pm$0.072&0.110$\pm$0.054&-0.023$\pm$0.055&0.153$\pm$0.042&0.098$\pm$0.081\\
\hline
hepatitis&\textbf{0.648$\pm$0.086}&0.618$\pm$0.094&0.366$\pm$0.133&0.370$\pm$0.122&0.495$\pm$0.130&0.407$\pm$0.120&0.249$\pm$0.153\\
\hline
hill-valley&-0.029$\pm$0.064&-0.029$\pm$0.064&0.064$\pm$0.034&0.050$\pm$0.041&0.089$\pm$0.051&0.071$\pm$0.023&\textbf{0.104$\pm$0.042}\\
\hline
horse-colic&0.674$\pm$0.086&0.669$\pm$0.074&0.691$\pm$0.046&\textbf{0.704$\pm$0.032}&0.664$\pm$0.087&0.684$\pm$0.096&0.630$\pm$0.088\\
\hline
ilpd-indian-liver&0.252$\pm$0.018&\textbf{0.269$\pm$0.043}&0.237$\pm$0.050&0.205$\pm$0.062&0.221$\pm$0.041&0.146$\pm$0.021&0.185$\pm$0.079\\
\hline
image-segmentation&0.921$\pm$0.025&0.921$\pm$0.019&\textbf{0.932$\pm$0.028}&0.893$\pm$0.051&0.615$\pm$0.073&0.916$\pm$0.029&0.899$\pm$0.061\\
\hline
ionosphere&\textbf{0.881$\pm$0.011}&0.868$\pm$0.011&0.866$\pm$0.029&0.866$\pm$0.043&0.804$\pm$0.047&0.818$\pm$0.048&0.836$\pm$0.029\\
\hline
iris&0.949$\pm$0.018&\textbf{0.959$\pm$0.041}&0.949$\pm$0.018&0.949$\pm$0.018&0.929$\pm$0.018&0.919$\pm$0.029&0.939$\pm$0.020\\
\hline
led-display&\textbf{0.725$\pm$0.013}&\textbf{0.725$\pm$0.013}&0.681$\pm$0.031&0.718$\pm$0.023&0.694$\pm$0.018&0.704$\pm$0.026&0.720$\pm$0.021\\
\hline
lenses&0.662$\pm$0.239&0.583$\pm$0.433&0.583$\pm$0.433&0.583$\pm$0.433&\textbf{0.762$\pm$0.274}&0.583$\pm$0.433&0.583$\pm$0.433\\
\hline
letter&0.953$\pm$0.001&0.939$\pm$0.002&\textbf{0.973$\pm$0.001}&0.966$\pm$0.002&0.348$\pm$0.018&0.964$\pm$0.002&0.964$\pm$0.001\\
\hline
libras&\textbf{0.848$\pm$0.020}&0.836$\pm$0.037&0.833$\pm$0.029&0.729$\pm$0.021&0.327$\pm$0.080&0.792$\pm$0.021&0.714$\pm$0.040\\
\hline
low-res-spect&0.829$\pm$0.037&0.812$\pm$0.022&0.857$\pm$0.022&\textbf{0.872$\pm$0.033}&0.684$\pm$0.037&0.860$\pm$0.028&0.866$\pm$0.034\\
\hline
lung-cancer&0.309$\pm$0.093&0.327$\pm$0.291&\textbf{0.360$\pm$0.151}&0.319$\pm$0.211&0.339$\pm$0.160&0.219$\pm$0.136&0.160$\pm$0.216\\
\hline
lymphography&\textbf{0.814$\pm$0.109}&0.804$\pm$0.124&0.631$\pm$0.067&0.748$\pm$0.092&0.509$\pm$0.065&0.721$\pm$0.113&0.778$\pm$0.123\\
\hline
magic&0.686$\pm$0.005&0.665$\pm$0.008&0.711$\pm$0.002&\textbf{0.727$\pm$0.005}&0.655$\pm$0.007&0.714$\pm$0.007&0.721$\pm$0.007\\
\hline
mammographic&0.663$\pm$0.006&\textbf{0.670$\pm$0.012}&0.572$\pm$0.025&0.632$\pm$0.028&0.593$\pm$0.052&0.584$\pm$0.026&0.632$\pm$0.023\\
\hline
miniboone&0.752$\pm$0.003&0.745$\pm$0.002&0.852$\pm$0.001&0.873$\pm$0.002&0.817$\pm$0.004&0.843$\pm$0.001&\textbf{0.875$\pm$0.001}\\
\hline
molec-biol-promoter&0.827$\pm$0.084&0.827$\pm$0.100&\textbf{0.865$\pm$0.084}&0.827$\pm$0.064&0.731$\pm$0.139&0.808$\pm$0.038&0.827$\pm$0.114\\
\hline
molec-biol-splice&0.714$\pm$0.025&0.698$\pm$0.030&0.926$\pm$0.020&0.937$\pm$0.004&0.888$\pm$0.006&0.922$\pm$0.014&\textbf{0.938$\pm$0.012}\\
\hline
monks-1&0.807$\pm$0.121&0.791$\pm$0.147&0.806$\pm$0.177&\textbf{0.935$\pm$0.079}&0.314$\pm$0.072&0.790$\pm$0.084&0.807$\pm$0.222\\
\hline
monks-2&0.558$\pm$0.086&\textbf{0.561$\pm$0.092}&0.431$\pm$0.151&0.496$\pm$0.128&0.000$\pm$0.000&0.188$\pm$0.120&0.173$\pm$0.156\\
\hline
monks-3&\textbf{0.917$\pm$0.055}&\textbf{0.917$\pm$0.055}&0.817$\pm$0.128&0.833$\pm$0.100&0.900$\pm$0.075&0.900$\pm$0.033&0.900$\pm$0.075\\
\hline
mushroom&\textbf{1.000$\pm$0.000}&\textbf{1.000$\pm$0.000}&\textbf{1.000$\pm$0.000}&\textbf{1.000$\pm$0.000}&\textbf{1.000$\pm$0.000}&\textbf{1.000$\pm$0.000}&\textbf{1.000$\pm$0.000}\\
\hline
musk-1&0.805$\pm$0.057&\textbf{0.810$\pm$0.061}&0.779$\pm$0.080&0.794$\pm$0.072&0.778$\pm$0.066&0.766$\pm$0.083&0.650$\pm$0.093\\
\hline
musk-2&0.919$\pm$0.008&0.914$\pm$0.006&0.946$\pm$0.005&\textbf{0.982$\pm$0.002}&0.968$\pm$0.005&0.915$\pm$0.004&0.970$\pm$0.006\\
\hline
nursery&0.956$\pm$0.005&0.945$\pm$0.005&0.996$\pm$0.001&\textbf{1.000$\pm$0.000}&0.742$\pm$0.006&0.995$\pm$0.001&\textbf{1.000$\pm$0.000}\\
\hline
oocytes\_merluccius\_nucleus\_4d&0.431$\pm$0.055&0.390$\pm$0.079&\textbf{0.562$\pm$0.074}&0.544$\pm$0.069&0.475$\pm$0.036&0.474$\pm$0.087&0.529$\pm$0.052\\
\hline
oocytes\_merluccius\_states\_2f&0.821$\pm$0.016&0.805$\pm$0.026&\textbf{0.836$\pm$0.025}&0.822$\pm$0.020&0.758$\pm$0.033&0.823$\pm$0.025&0.827$\pm$0.021\\
\hline
oocytes\_trisopterus\_nucleus\_2f&0.595$\pm$0.032&0.559$\pm$0.027&\textbf{0.639$\pm$0.033}&0.605$\pm$0.025&0.550$\pm$0.025&0.619$\pm$0.028&0.619$\pm$0.038\\
\hline
oocytes\_trisopterus\_states\_5b&0.840$\pm$0.032&0.824$\pm$0.022&0.839$\pm$0.021&0.857$\pm$0.031&0.613$\pm$0.031&0.839$\pm$0.017&\textbf{0.864$\pm$0.013}\\
\hline
optical&0.973$\pm$0.006&0.959$\pm$0.006&\textbf{0.983$\pm$0.002}&0.981$\pm$0.004&0.872$\pm$0.012&0.981$\pm$0.004&0.974$\pm$0.002\\
\hline
ozone&\textbf{0.256$\pm$0.050}&0.045$\pm$0.080&-0.001$\pm$0.001&0.025$\pm$0.045&0.000$\pm$0.000&-0.001$\pm$0.001&0.213$\pm$0.056\\
\hline
page-blocks&0.824$\pm$0.020&0.795$\pm$0.021&0.848$\pm$0.021&0.852$\pm$0.023&0.524$\pm$0.083&0.845$\pm$0.023&\textbf{0.856$\pm$0.024}\\
\hline
parkinsons&0.767$\pm$0.117&0.741$\pm$0.121&\textbf{0.786$\pm$0.069}&0.721$\pm$0.085&0.659$\pm$0.171&0.692$\pm$0.097&0.752$\pm$0.056\\
\hline
pendigits&0.990$\pm$0.002&0.990$\pm$0.002&\textbf{0.994$\pm$0.002}&0.992$\pm$0.001&0.770$\pm$0.008&0.989$\pm$0.001&0.990$\pm$0.002\\
\hline
pima&0.448$\pm$0.030&\textbf{0.458$\pm$0.026}&0.427$\pm$0.031&0.423$\pm$0.044&0.383$\pm$0.012&0.451$\pm$0.060&0.437$\pm$0.042\\
\hline
pittsburg-bridges-MATERIAL&0.846$\pm$0.051&0.828$\pm$0.070&\textbf{0.849$\pm$0.077}&0.736$\pm$0.098&0.721$\pm$0.051&0.735$\pm$0.085&0.695$\pm$0.056\\
\hline
pittsburg-bridges-REL-L&0.611$\pm$0.047&\textbf{0.626$\pm$0.084}&0.573$\pm$0.083&0.387$\pm$0.147&0.505$\pm$0.049&0.456$\pm$0.110&0.414$\pm$0.119\\
\hline
pittsburg-bridges-SPAN&\textbf{0.534$\pm$0.164}&0.512$\pm$0.178&0.435$\pm$0.131&0.445$\pm$0.088&0.282$\pm$0.051&0.348$\pm$0.081&0.366$\pm$0.115\\
\hline
pittsburg-bridges-T-OR-D&0.266$\pm$0.249&0.282$\pm$0.340&0.296$\pm$0.346&\textbf{0.503$\pm$0.212}&0.234$\pm$0.234&0.318$\pm$0.191&0.356$\pm$0.229\\
\hline
pittsburg-bridges-TYPE&0.541$\pm$0.100&0.541$\pm$0.100&\textbf{0.565$\pm$0.122}&0.483$\pm$0.117&0.249$\pm$0.098&0.539$\pm$0.131&0.437$\pm$0.075\\
\hline
planning&\textbf{0.104$\pm$0.073}&0.082$\pm$0.089&0.024$\pm$0.043&-0.020$\pm$0.046&0.000$\pm$0.000&0.001$\pm$0.089&-0.098$\pm$0.036\\
\hline
plant-margin&0.841$\pm$0.017&0.816$\pm$0.008&\textbf{0.885$\pm$0.007}&0.708$\pm$0.010&0.360$\pm$0.030&0.859$\pm$0.007&0.711$\pm$0.006\\
\hline
plant-shape&0.655$\pm$0.013&0.587$\pm$0.013&\textbf{0.665$\pm$0.011}&0.456$\pm$0.017&0.192$\pm$0.014&0.642$\pm$0.018&0.533$\pm$0.029\\
\hline
plant-texture&0.811$\pm$0.008&0.788$\pm$0.007&\textbf{0.846$\pm$0.007}&0.516$\pm$0.300&0.407$\pm$0.021&0.838$\pm$0.011&0.718$\pm$0.012\\
\hline
post-operative&-0.069$\pm$0.193&\textbf{-0.032$\pm$0.238}&-0.115$\pm$0.126&-0.215$\pm$0.161&-0.132$\pm$0.099&-0.091$\pm$0.114&-0.081$\pm$0.171\\
\hline
ringnorm&\textbf{0.968$\pm$0.002}&\textbf{0.968$\pm$0.002}&0.965$\pm$0.003&0.958$\pm$0.008&0.962$\pm$0.005&0.918$\pm$0.006&0.958$\pm$0.005\\
\hline
seeds&0.913$\pm$0.054&0.899$\pm$0.052&\textbf{0.942$\pm$0.046}&0.906$\pm$0.043&0.492$\pm$0.012&0.913$\pm$0.035&0.906$\pm$0.043\\
\hline
semeion&0.939$\pm$0.017&0.937$\pm$0.015&0.948$\pm$0.016&\textbf{0.951$\pm$0.013}&0.768$\pm$0.022&0.947$\pm$0.017&0.916$\pm$0.015\\
\hline
soybean&0.924$\pm$0.012&0.920$\pm$0.016&\textbf{0.942$\pm$0.023}&0.905$\pm$0.030&0.722$\pm$0.062&0.927$\pm$0.023&0.916$\pm$0.026\\
\hline
spambase&0.891$\pm$0.008&0.886$\pm$0.004&0.908$\pm$0.007&\textbf{0.910$\pm$0.002}&0.891$\pm$0.009&0.906$\pm$0.003&0.906$\pm$0.010\\
\hline
spect&0.267$\pm$0.067&0.295$\pm$0.312&0.081$\pm$0.150&0.079$\pm$0.176&\textbf{0.376$\pm$0.095}&0.246$\pm$0.159&0.083$\pm$0.136\\
\hline
spectf&0.600$\pm$0.100&\textbf{0.675$\pm$0.109}&0.550$\pm$0.087&0.300$\pm$0.187&0.400$\pm$0.122&0.450$\pm$0.112&0.500$\pm$0.158\\
\hline
statlog-australian-credit&0.171$\pm$0.094&\textbf{0.180$\pm$0.050}&0.153$\pm$0.043&0.162$\pm$0.049&-0.005$\pm$0.018&0.122$\pm$0.055&0.164$\pm$0.084\\
\hline
statlog-german-credit&0.422$\pm$0.041&0.411$\pm$0.036&0.437$\pm$0.014&0.435$\pm$0.051&0.377$\pm$0.018&\textbf{0.446$\pm$0.027}&0.436$\pm$0.025\\
\hline
statlog-heart&0.748$\pm$0.034&\textbf{0.764$\pm$0.026}&0.672$\pm$0.053&0.696$\pm$0.057&0.718$\pm$0.063&0.726$\pm$0.089&0.598$\pm$0.053\\
\hline
statlog-image&0.972$\pm$0.006&0.964$\pm$0.008&0.984$\pm$0.006&0.980$\pm$0.004&0.826$\pm$0.033&0.977$\pm$0.006&\textbf{0.985$\pm$0.003}\\
\hline
statlog-landsat&0.876$\pm$0.005&0.872$\pm$0.007&0.879$\pm$0.005&0.878$\pm$0.005&0.629$\pm$0.041&0.877$\pm$0.009&\textbf{0.886$\pm$0.003}\\
\hline
statlog-shuttle&\textbf{0.999$\pm$0.000}&0.998$\pm$0.000&\textbf{0.999$\pm$0.000}&\textbf{0.999$\pm$0.000}&0.997$\pm$0.001&\textbf{0.999$\pm$0.000}&\textbf{0.999$\pm$0.000}\\
\hline
statlog-vehicle&0.639$\pm$0.016&0.637$\pm$0.019&0.659$\pm$0.024&0.687$\pm$0.030&0.475$\pm$0.027&0.671$\pm$0.018&\textbf{0.706$\pm$0.015}\\
\hline
steel-plates&0.710$\pm$0.012&0.703$\pm$0.006&\textbf{0.751$\pm$0.011}&0.740$\pm$0.016&0.441$\pm$0.053&0.724$\pm$0.008&0.738$\pm$0.009\\
\hline
synthetic-control&0.984$\pm$0.008&0.978$\pm$0.016&0.986$\pm$0.007&\textbf{0.988$\pm$0.007}&0.600$\pm$0.038&0.984$\pm$0.006&0.960$\pm$0.013\\
\hline
teaching&0.517$\pm$0.050&\textbf{0.527$\pm$0.077}&0.478$\pm$0.088&0.489$\pm$0.106&0.361$\pm$0.071&0.517$\pm$0.069&0.478$\pm$0.075\\
\hline
thyroid&0.691$\pm$0.036&0.694$\pm$0.026&0.954$\pm$0.022&0.977$\pm$0.010&0.951$\pm$0.006&\textbf{0.989$\pm$0.004}&0.987$\pm$0.006\\
\hline
tic-tac-toe&0.958$\pm$0.008&0.951$\pm$0.008&0.977$\pm$0.008&0.974$\pm$0.012&0.944$\pm$0.015&\textbf{0.979$\pm$0.012}&0.972$\pm$0.012\\
\hline
titanic&0.445$\pm$0.029&0.445$\pm$0.029&0.427$\pm$0.008&0.427$\pm$0.008&\textbf{0.453$\pm$0.003}&0.427$\pm$0.008&0.427$\pm$0.008\\
\hline
twonorm&0.959$\pm$0.008&\textbf{0.960$\pm$0.007}&0.957$\pm$0.004&0.948$\pm$0.004&0.949$\pm$0.007&0.949$\pm$0.006&0.950$\pm$0.005\\
\hline
vertebral-column-2clases&\textbf{0.653$\pm$0.037}&0.650$\pm$0.039&0.651$\pm$0.082&0.539$\pm$0.122&0.573$\pm$0.086&0.572$\pm$0.083&0.562$\pm$0.093\\
\hline
vertebral-column-3clases&0.762$\pm$0.025&\textbf{0.767$\pm$0.018}&0.738$\pm$0.056&0.691$\pm$0.059&0.540$\pm$0.155&0.740$\pm$0.062&0.734$\pm$0.060\\
\hline
wall-following&0.924$\pm$0.007&0.919$\pm$0.005&0.977$\pm$0.006&\textbf{0.997$\pm$0.002}&0.919$\pm$0.021&0.994$\pm$0.001&0.995$\pm$0.002\\
\hline
waveform&\textbf{0.808$\pm$0.012}&0.798$\pm$0.021&0.786$\pm$0.020&0.779$\pm$0.011&0.765$\pm$0.023&0.771$\pm$0.013&0.769$\pm$0.014\\
\hline
waveform-noise&0.776$\pm$0.010&0.775$\pm$0.011&\textbf{0.803$\pm$0.009}&0.795$\pm$0.008&0.752$\pm$0.009&0.794$\pm$0.014&0.785$\pm$0.013\\
\hline
wine&\textbf{0.991$\pm$0.015}&\textbf{0.991$\pm$0.015}&\textbf{0.991$\pm$0.015}&\textbf{0.991$\pm$0.015}&0.904$\pm$0.046&0.974$\pm$0.028&0.974$\pm$0.029\\
\hline
wine-quality-red&\textbf{0.518$\pm$0.016}&0.512$\pm$0.021&0.492$\pm$0.031&0.419$\pm$0.022&0.255$\pm$0.007&0.494$\pm$0.026&0.434$\pm$0.016\\
\hline
wine-quality-white&\textbf{0.532$\pm$0.012}&0.529$\pm$0.015&0.523$\pm$0.011&0.512$\pm$0.006&0.090$\pm$0.020&0.511$\pm$0.008&0.502$\pm$0.006\\
\hline
yeast&0.508$\pm$0.042&0.502$\pm$0.029&0.485$\pm$0.034&0.496$\pm$0.020&0.189$\pm$0.055&\textbf{0.519$\pm$0.027}&0.505$\pm$0.027\\
\hline
zoo&\textbf{0.986$\pm$0.024}&\textbf{0.986$\pm$0.024}&\textbf{0.986$\pm$0.024}&\textbf{0.986$\pm$0.024}&0.918$\pm$0.061&\textbf{0.986$\pm$0.024}&\textbf{0.986$\pm$0.024}\\
\hline
\hline
AVERAGE&\textbf{0.685$\pm$0.043}&0.675$\pm$0.051&0.673$\pm$0.047&0.663$\pm$0.055&0.550$\pm$0.046&0.663$\pm$0.047&0.660$\pm$0.052\\
\hline
\caption{The performances of methods is compared on 115 UCI datasets using Cohen's kappa coefficient.}
\label{kappaUCI}
\end{longtable}
\end{center}
\end{landscape}

\bibliography{main}

\section*{Competing interests}
The authors declare that they have no known competing interests.

\section*{CRediT authorship contribution statement}
\textbf{Prashant Gupta}: Design of this study, Methodology, Software, Investigation. \textbf{Aashi Jindal}: Design of this study, Methodology, Software. \textbf{Jayadeva}: Conceptualization of this study, Design. \textbf{Debarka Sengupta}: Conceptualization of this study, Design. \textbf{Suresh Chandra}: Conceptualization of this study, Methodology.

\end{document}